\documentclass[preprint,i12pt,authoryear]{elsarticle}

\usepackage{amssymb}
\usepackage{amsmath}
\usepackage{algorithm}
\usepackage{algpseudocode}

\usepackage{subcaption}
\usepackage{placeins}
\usepackage{algorithm}
\usepackage{algpseudocode}
\usepackage{todonotes}
\usepackage[colorlinks,
            citecolor=blue,
            urlcolor=blue,
            linkcolor=blue]{hyperref}
\usepackage{cleveref}

\usepackage{tikz}
\usepackage{graphicx}
\usepackage{bbm}
\usepackage{mathtools}
\journal{tbd}

\DeclareMathAlphabet{\mathcall}{OMS}{cmsy}{m}{n}
\newcommand{\secref}[1]{\hyperref[#1]{Section~\ref*{#1}}}
\newcommand{\theref}[1]{\hyperref[#1]{Theorem~\ref*{#1}}}
\newcommand{\lemref}[1]{\hyperref[#1]{Lemma~\ref*{#1}}}
\newcommand{\figref}[1]{\hyperref[#1]{Figure~\ref*{#1}}}
\newcommand{\figsref}[1]{Figures\hyperref[#1]{~\ref*{#1}}}
\newcommand{\tabref}[1]{\hyperref[#1]{Table~\ref*{#1}}}

 \newcommand{\R}{\mathbb{R}}

\begin{document}
\newcommand{\JR}[1]{\todo[author=JR,color=orange!50,size=\small]{#1}}
\newcommand{\JRil}[1]{\todo[inline,author=JR,color=orange!50,size=\small]{#1}}
\newcommand{\FLil}[1]{\todo[inline,author=FL,color=yellow!50,size=\small]{#1}}
\newcommand{\Ical}{\mathcal{I}}
\newcommand{\AGil}[1]{\todo[inline,author=AG,color=green!50,size=\small]{#1}}
\newcommand{\BAZil}[1]{\todo[inline,author=BAZ,color=violet!50,size=\small]{#1}}

\begin{frontmatter}



\affiliation[1]{organization={Department of Data Science, Friedrich-Alexander-Universität Erlangen-Nürnberg Erlangen-Nürnberg},
postcode={91052},
city={Erlangen},
country={Germany}}

\affiliation[2]{organization={Institute of Micro- and Nanostructure Research (IMN) $\&$ Center for Nanoanalysis and Electron Microscopy (CENEM), Department of Materials Science and Engineering, Friedrich-Alexander-Universität Erlangen-Nürnberg},
postcode={91058},
city={Erlangen},
country={Germany}}

\affiliation[3]{organization={Department of Mathematics, Linköping University},
postcode={SE-581 83},
city={Linköping},
country={Sweden}}

\affiliation[4]{organization={Electric Power Systems unit, RISE - Research Institutes of Sweden},
postcode={SE-114 28},
city={Stockholm},
country={Sweden}}

\author[1]{Anuraag Mishra}
\ead{anuraag.mishra@fau.de}
\ead[url]{https://github.com/mishraanuraagx}

\author[1]{Andrea Gilch}
\ead{andrea.gilch@fau.de}
\ead[url]{
https://www.datascience.nat.fau.eu/research/groups/ouda/members/andrea-gilch/}

\author[2]{Benjamin Apeleo Zubiri\corref{cor}}
\ead{benjamin.apeleo.zubiri@fau.de}
\ead[url]{https://www.em.tf.fau.de/person/benjamin-apeleo-zubiri-geb-winter/}

\author[3,4]{Jan Rolfes\corref{cor}}
\ead{jan.rolfes@liu.se}
\ead[url]{https://liu.se/en/employee/janro69}
\cortext[cor]{corresponding author} 

\author[1]{Frauke Liers\corref{cor}}
\ead{frauke.liers@fau.de}
\ead[url]{https://www.datascience.nat.fau.eu/research/groups/ouda/members/frauke-liers/}

\title{
High-Quality Tomographic Image Reconstruction Integrating Neural Networks and Mathematical Optimization 
}




\begin{abstract}

In this work, we develop a novel technique for reconstructing images from projection-based nano- and microtomography.
Our contribution focuses on enhancing reconstruction quality, particularly for specimen composed of homogeneous material phases connected by sharp edges. This is accomplished by training a neural network to identify edges within subpictures. The trained network is then integrated into a mathematical optimization model, to reduce artifacts from previous reconstructions. To this end, the optimization approach favors solutions according to the learned predictions, however may also determine alternative solutions if these are strongly supported by the raw data. Hence, our technique successfully incorporates knowledge about the homogeneity and presence of sharp edges in the sample and thereby eliminates blurriness. Our results on experimental datasets show significant enhancements in interface sharpness and material homogeneity compared to benchmark algorithms. Thus, our technique produces high-quality reconstructions, showcasing its potential for advancing tomographic imaging techniques.

\end{abstract}


\end{frontmatter}

\section{Introduction}\label{sec:Introduction}


The primary objective of projection-based nano- and microtomography is to extract three-dimensional information from a series of two-dimensional projections of a \emph{sample} or \emph{specimen} spanning a scale from nanometers to millimeters  \cite{burnett2019completing,withers2021x,Goetz2025}. Reconstructions can be generated using techniques such as electron tomography (ET), nano-computed X-ray tomography (nano-CT), and micro-computed X-ray tomography (micro-CT), depending on the required sample size and spatial resolution. Despite significant advancements and optimizations in tomographic reconstruction techniques over the past decades, developing high-quality, problem-specific reconstruction methods remains a relevant and challenging area of research \cite{Saiprasad}. Typically, these reconstruction problems are addressed using mathematical optimization or machine learning methods, which aid in obtaining a reconstructed image with high \emph{fidelity}, i.e., ensuring that the reconstructed object closely resembles the ground truth. In this work, we present a hybrid algorithm that incorporates knowledge gained from data within a mathematical optimization framework, demonstrating that it achieves high-quality reconstructions in a reasonable time frame.

In transmission electron microscopes and X-ray microscopes, the sample holder or stage is positioned between the stationary X-ray or electron beam source and the detector array. The measured intensities from the beam that passes through the sample correspond to specific properties of the sample, such as local density (i.e., mass attenuation) and thickness, allowing for the generation of interpretable contrast. In this context, the \emph{high-angle annular dark-field} 
scanning transmission electron microscopy 
imaging mode has become a standard technique for ET in materials science \cite{Leary2019}. Nano- and micro-CT measurements can be conducted in \emph{absorption contrast mode}, producing mass-thickness contrast in accordance with Beer-Lambert's law \cite{withers2021x}. In any measurement, the sample holder is tilted to capture projections from multiple angles. These projections are then utilized for the three-dimensional (3D) reconstruction of the specimen, ideally over a 180° tilt-angle range, which is crucial for achieving an accurate reconstruction \cite{Leary2019}. 

In an idealized setting, the projections of an $n$-dimensional picture on an $m$-dimensional space (with $m\leq n$) can be modeled as a noiseless system of linear equations of the form
    \begin{align}
        Rf = p \label{Rf=p}.
    \end{align}
Here, the vector $f \in \mathbb{R}^n$ denotes the brightness of each of the $n$ many pixels, $p \in \mathbb{R}^m$ the $m$-dimensional projection data. The matrix $R \in \mathbb{R}^{m \times n}$ is a discretized Radon transform. An analytical solution to the above system can be obtained by \emph{filtered backprojection} 
and was initially analyzed in \cite{kak2001principles}. This approach can model various projection geometries, including parallel, fan, and cone beam configurations. However,  \eqref{Rf=p} represents an idealized scenario that neglects uncertainties in the measurements and the potentially limited number of available projections.  Since uncertainties in the data are often unavoidable -- due to factors such as noise, restricted beam time, high costs associated with acquiring the projections, and the beam sensitivity of the material (which can only withstand a certain amount of X-ray or electron dose) -- the reconstruction model presented in \eqref{Rf=p} may not yield any feasible solution $f$. As a result, the goal is to find an image $f$ that minimizes the distance between $Rf$ and the projection data $p$. In mathematical terms, the image reconstruction problem can be expressed as follows
    \begin{align}
        \min\limits_{f} \, \lVert Rf - p\rVert_2^2 \label{minlimited},
    \end{align}

This setting is often referred to as an \emph{inconsistent scenario} since only $m \ll n$ projections are acquired leading to an underdetermined linear equation system \eqref{Rf=p} and, thus, to undersampled information. 

Broadly speaking, established reconstruction techniques for inconsistent scenarios are either based on iterative techniques that utilize gradient information such as e.g., the simultaneous algebraic reconstruction technique (SART) 
\cite{slaney1988principles} and the simultaneous iterative reconstruction technique (SIRT) \cite{gilbert1972iterative,kak2001principles}, global optimization techniques such as Compressed Sensing (CS) \cite{candes2006stable} and a combination of the two approaches such as TVR-DART \cite{zhuge2015tvr}. 

Whereas SART and SIRT target \eqref{minlimited} directly, the classical CS signal processing technique is able to achieve improved reconstructions from undersampled information by utilizing prior knowledge about the signal to reconstruct. In particular, if sparsity of the signal can be assumed, the reconstruction can improve drastically, see e.g., the seminal paper 
\cite{candes2006stable}. Applications of CS include various areas, we refer to \cite{qaisar2013compressive} for an overview. Moreover, CS is often used in a way to steer solutions towards discrete gray values, specifically by incorporating various regularization terms such as high order total variation (HOTV) \cite{xi2020study}, wavelets \cite{lustig2007sparse}, total generalized variation (TGV) \cite{knoll2011second} or total variation (TV) regularization \cite{leary2013compressed}. The source \cite{leary2013compressed} serves as a valuable comparison as it provides numerical results specifically for nanotomography and solves the resulting modified CS fairly efficiently with a conjugate gradient descent algorithm.

The benchmark problem considered here consists of an objective function \eqref{minlimited}, which is extended by an additional term that prefers a small total variation (TV)-norm of the gradient of $f$. The TV-norm $\| f \|_{TV}$ for a picture $f$ is defined via $\| f \|_{TV}:= \sum_{j=1}^n \| \nabla f_j \|_1$, where the norm $\|\cdot \|_1$ as usual sums over the corresponding absolute values. The additional term in the objective penalizes large TV-norm values and is of form $\lambda \|f\|_{TV}$, where the scalar $\lambda\geq 0$ is an appropriately chosen parameter:   
\begin{align}
    \min\limits_{f} \, \lVert Rf - p\rVert_2^2 + \lambda \|f\|_{TV}\label{TVRdartmodel}.
\end{align}

As a result, Compressed Sensing (CS) aims to guide the reconstructed solution towards discrete gray values, yielding impressive results in the tomography of homogeneous materials. Therefore, alongside SIRT, CS serves as our second benchmark algorithm.

In addition to the aforementioned methods, we would like to emphasize the total variation regularized discrete algebraic reconstruction technique (TVR-DART) \cite{zhuge2015tvr} as an exemplary iterative technique designed specifically for samples with only a few distinct density values. For this reason, it serves as our third benchmark algorithm. Essentially, TVR-DART incorporates additional information into \eqref{TVRdartmodel} and seeks local optima for the resulting non-convex problem. While this approach often leads to effective performance, the authors in \cite{zhuge2015tvr} acknowledge that it can sometimes result in divergence under certain conditions.

As a fourth benchmark, we compare our results with the \emph{compressed sensing for homogeneous
materials} (CSHM) algorithm introduced in \cite{Kreuz}. This algorithm is an extension of the classical Compressed Sensing (CS) technique, as described in \cite{candes2006stable}.  
Specifically, the authors in \cite{Kreuz} focus additionally on samples in which a relatively homogeneous sample consisting of piece-wise constant regions is assumed. It is worth noting that although CSHM has effectively addressed the homogeneous characteristics of specific samples, the fact that many tomography samples are surrounded by air or vacuum and fit entirely within the imaging field of view, some inconsistencies in the acquired raw data persist. These include issues such as low signal-to-noise ratios, crystallographic Bragg contrast, beam broadening in thicker samples, and inaccuracies in tilt series alignment. Such inconsistencies can result in blurred edges in the 3D reconstruction and unwanted contrast variations in materials that are originally homogeneous. 

\subsection{Our Contribution}

Building upon CSHM, our contribution is to address these challenges as follows. Overall, we incorporate information about the reconstruction of sharp edges. To this end, we train a neural network that indicates whether the raw data on a $3\times 3$ subpicture supports the presence of a sharp edge between homogeneous material phases. The resulting function is then integrated into a mathematical optimization model, that aims to remove blurry edges present in a CSHM reconstruction as a post-processing step.
As neural networks may not generalize reliably to unseen data, our optimization approach prioritizes solutions consistent with the learned predictions, but also considers alternative solutions when these yield superior quality.
The computational results obtained for our tomography specimen show that this post-processing improves the quality of CSHM reconstructions significantly and can be obtained within reasonable computing time. Furthermore, our approach allows for significant parallelization with respect to the considered subpictures, which creates the potential to reduce the computing time of our method even further.  

\subsection{Paper Structure}

We first survey the literature on incorporating physics-driven information either via model-based approaches or via learning-based approaches. Subsequently, we emphasize the incorporation of data-driven information into \emph{Mixed-Integer Programs} (MIP), which is described in \secref{sec:related_work}. The mathematical reconstruction scheme of CSHM as well as related literature is described in \secref{subsec:cshm}. As conic programming is at the core of CSHM, we therefore have a natural base approach, which is going to be defined in \secref{sec:framework_integrated} and refined in \secref{sec:framework_reopt} to significantly reduce the computational burden and allow for parallelization without significant quality losses. 
We continue by providing experimental details in \secref{sec:exp-details} such as the considered ground truths and the considered training data for the neural network and close with the computational results in \secref{sec:compresults}. Here, we compare the approaches introduced in \secref{sec:methods} to our benchmark algorithms. 
We demonstrate that our novel approach 
leads to high-quality reconstructions within reasonable time. 
Lastly, we discuss these results and some future extensions 
as well as future applications in \secref{sec:conclusion}.

\section{Related Work}\label{sec:related_work}

The primary motivation for the approach presented in this article is to incorporate as much prior knowledge about the observed sample as possible, whether this information is derived from modeling the sample's physical properties or from data-driven insights. Apart from CSHM, which extends Compressed Sensing with model-based information, we would like to highlight the work on Model-Based Iterative Reconstruction (MBIR) Techniques, see e.g., \cite{venkatakrishnan2014model} and the references therein. MBIR algorithms in general aim to incorporate model-based information into iterative algorithms such as TVR-DART and SIRT by modeling imaging physics as additional constraints. However, to the best of our knowledge, the current literature mostly focuses on specific anomalies such as Bragg contrast in \cite{venkatakrishnan2014model}. In contrast, we do not focus on specific anomalies, but we rather present a more general approach to address a broader range of inconsistencies in the tomographic experiment, such as tilt series alignment inaccuracies, non-linear contrast contributions, or noise. 

Utilizing deep learning in order to enhance image reconstructions has particularly gained traction since the mid 2010s. We refer the reader to the comprehensive overview by \cite{Wang2020Deep}, which discusses the usage of deep neural networks for tomographic reconstruction, including representative techniques and open challenges. This research area can be broadly categorized into three main approaches: direct end-to-end reconstruction, hybrid integration with classical models, and post-processing enhancement on which we will briefly elaborate:

Direct end-to-end methods use neural networks to map projection data directly to reconstructed images without explicitly incorporating the forward model \eqref{Rf=p}, see e.g. \cite{Han2018}, \cite{Zhu2018}, \cite{Ye2018}. While powerful, these approaches typically require large, diverse training datasets and may lack on theoretical justification especially for low-data scenarios.
Hybrid methods, on the other hand, combine deep neural networks with iterative reconstruction schemes, e.g. by learning well-performing iterative steps as in \cite{Adler} or \cite{Chen2020AirNet}, or by incorporating neural networks either as a pretrained \cite{Chen2018} or adaptively trained \cite{Baguer_2020} prior into an iterative reconstruction method. These approaches balance physical modeling with data-driven regularization but often at high training complexity. 

Post-processing approaches apply trained networks in the image domain after conventional reconstruction, acting as denoisers \cite{Kang2017} or artifact suppressors \cite{Christiansen2018} to compensate for systematic weaknesses of classical approaches. Moreover, these approaches encompass feature identification in reconstructed images such as fluorescent labels in unlabeled images \cite{Christiansen2018}.

Thereby, our work addresses a gap in the literature by bridging data-driven operator learning with exact mathematical modeling, opening up novel hybrid strategies for image reconstruction. 
Typically, neural networks are trained 
such that some appropriate loss function is minimized, i.e., a mathematical optimization problem is solved. Difficult, non-convex optimization problems arise for some of the standard activation functions that are often solved locally by gradient-based first order algorithms. Although this often leads to good results, such gradient-based approaches typically get stuck in local minima that may be far away from best possible, globally optimal solutions. In general, no guarantee about solution quality can be given. 

In contrast, it has been observed by different authors~\cite{fischetti2017deep, lecun98, tjeng2017} that neural network training using the specific rectified linear unit (ReLU) activation function can be modeled as a mixed-integer optimization problem (MIP), which can be linearized by standard approaches. Although solving an MIP is in general an NP-hard optimization problems, i.e., difficult both in theory and in practice, modern available MIP solvers can solve even huge instances to global optimality within reasonable time. In our mathematical optimization approach, we use the model~\cite{fischetti2017deep} to represent a deep neural network (DNN) by mixed-integer constraints that can easily be integrated in a MIP solver as a subproblem. Details will be given in subsequent sections.

\section{Enhanced Algorithms Incorporating Data-Driven Information}\label{sec:methods}
\subsection{Big Picture}\label{sec:big_picture} 

In general, tomographic reconstruction methods provide good results but may struggle to preserve uniform pixel intensity regions and sharp edge features,
i.e.,  
a rapid change of pixel values within small image subregions. 
Here, this is captured by the concept of edge intensity, which is generally defined as the variation in pixel values within a small image subregion. 
An image contains sharp edges when the edge intensities in its subregions are either small (indicating constant pixel values) or reach its upper bound (indicating an abrupt change of pixel values).
Various methods for measuring edge intensity are available and will be discussed in Chapter \ref{subsec:NN_Design}.

\begin{figure}[htbp]
    \centering
    \includegraphics[width = \linewidth]{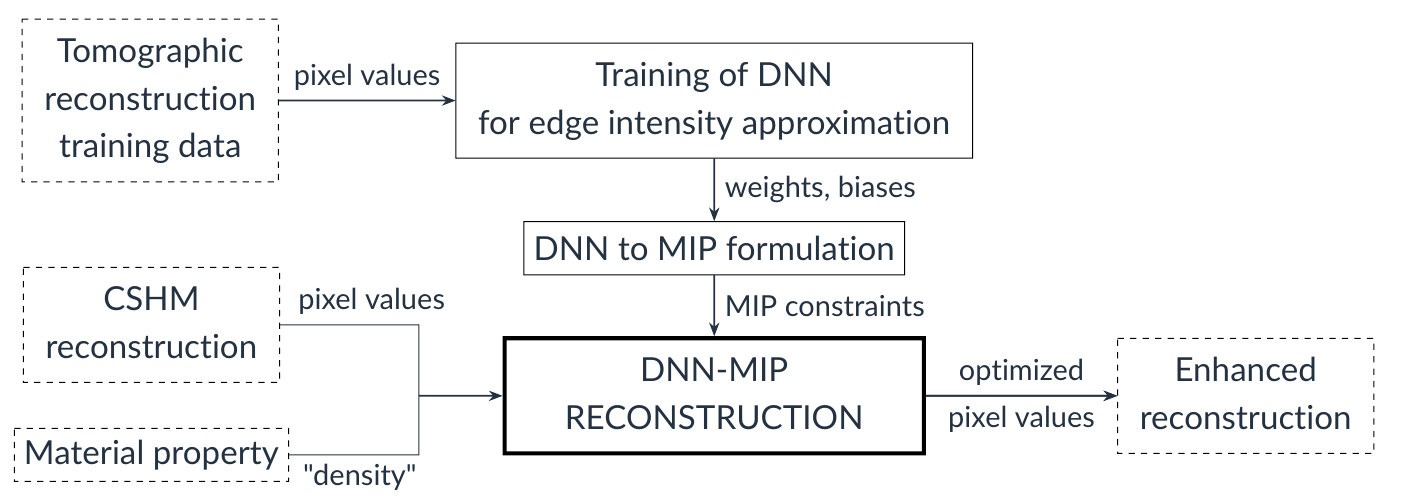}
    \caption{Overall process flow of the DNN-MIP reconstruction model}
    \label{fig:flowchart}
\end{figure}

We will next give a high-level explanation of our \emph{DNN-MIP reconstruction} model. \figref{fig:flowchart} illustrates the overall process flow. The model enhances image reconstructions by integrating learned edge information into the CSHM optimization model which is explained in Section \ref{subsec:cshm}, see also the left column in \figref{fig:flowchart}. This methodology leverages the strengths of deep learning for edge prediction while maintaining the formal rigor and tractability of mathematical optimization. Therefore, it is essential to formulate all components in mathematical terms, to be able to extend CSHM. 

\FloatBarrier

We aim to reconstruct an image with enhanced edge sharpness by optimizing pixel values so that the edge intensity within each subregion is either near its upper bound (denoting an edge) or zero (denoting homogeneity). 
In the first step (\figref{fig:flowchart}, top), we train the DNN for 
measuring the edge intensity of images, 
following standard procedures as described, for example, in \cite{goodfellow2016deep}. The DNN is fed with pixel values from image training data and approximates the edge intensity.
Subsequently, see box "DNN to MIP formulation", we extract the learned hyper parameters (weights and biases of all nodes) and set up the corresponding MIP defined by \eqref{eq:DNN} from upcoming Section \ref{subsec:Model}.  
We thus obtain an operator, defined by MIP constraints, for the mathematical calculation of the DNN's output, that is, the approximated edge intensity for a given input image. 
We extend the CSHM model, see box "CSHM reconstruction",  by the DNN-operator and a corresponding objective function which is penalizing intermediate values of the edge intensity. As a result, the optimization framework gives preference to solutions close to the learned predictions, yet it also identifies alternative ones whenever they offer superior outcomes.
Consequently, the integrated optimization problem promotes edge sharpness alongside accurate image reconstruction.
Furthermore, we extend the model with material property information (see bottom left) by assuming a constant pixel density for the homogeneous material. To this end, we obtain the final DNN-MIP reconstruction model realized in a multi-objective optimization problem balancing between preserving characteristics of the reconstructed image, satisfying the physical material properties and promoting edge sharpness.

\subsection{Compressed Sensing for Homogeneous Material: CSHM Reconstruction}
\label{subsec:cshm}
\cite{Kreuz} have introduced a mathematical optimization model for compressed sensing in case the underlying material is homogeneous. It extends~\eqref{TVRdartmodel} by additional constraints and an additional term in the objective function. In order to be self-contained, we repeat this model here. The extensions are two-fold. Firstly, it is taken into account that a homogeneous material has a constant pixel density, denoted by $\omega\in\R_{\geq 0}$. This density value can typically be estimated beforehand. A pixel value should either be small (in case it represents no material) or close to $\omega$ (in case it represents material). We now add for each pixel $j\in \{1,\ldots,n\}$ a variable $d_j\geq 0$ and a constraint $d_j\geq f_j-\omega$. If $d_j$ takes value zero, this constraint models that $f_j$ at most has value $\omega$. As the value of $\omega$ is based on estimation, we penalize quadratically large values of $d_j$ in the objective. We do this by adding a term of form $\mu\|d\|^2_2$, where again $\mu\geq 0$ is an appropriately chosen parameter. 

Secondly, another set of constraints is added in the model as follows. The ideal reconstruction \eqref{Rf=p} equation systems reads 

$$p_i =\sum_{j=1}^n R_{ij}f_j,$$
where $p\geq 0, R\geq 0, f\geq 0$. As a result, if a pixel $j$ in picture $f$ gets passed by a projection ray $i$, then $R_{ij}>0$, and it holds that $R_{ij}f_j\leq \sum_{j=1}^n R_{ij}f_j = p_i$. Thus, for each pixel $1\leq j\leq n$, the following constraint needs to be satisfied 

$$f_j\leq \min_{1\leq i\leq m:R_{ij}\not=0} \frac{p_i}{R_{ij}}.$$  The CSHM model now reads as

\begin{subequations}
\label{prob:cshm}
\begin{align}
    \min_{f}\  CSHM(f) = \min_{f}\ & \| R f - p \|_2^2 + \lambda \| f \|_{TV} + \mu \| d \|_2^2 \label{CSHM_obj_function}\\
& d_j \geq f_j - \omega && \forall 1 \leq j \leq n,\label{constr:CSHM1}\\
& d_j \geq 0 && \forall 1 \leq j \leq n,\\
& f_j \leq \min_{1\leq i\leq m:R_{ij}\not=0} \frac{p_i}{R_{ij}} && \forall 1 \leq j \leq n,\\
& 
f_j \geq 0 && \forall 1 \leq j \leq n.\label{constr:CSHM4}
\end{align}
\end{subequations}

In \cite{Kreuz}, the benchmark algorithms from Section \ref{sec:Introduction} were evaluated, where for homogeneous materials it turned out that CSHM yields best results. In this work, we further extend the CSHM optimization model by integrating additional knowledge about material edges that is learned from data. 

\subsection{DNN Design}
\label{subsec:NN_Design}
We follow the flowchart from Section \ref{sec:big_picture} and, first, set up a DNN to obtain the edge intensity of an image subregion. Therefore, we select an appropriate edge detector, the Sobel operator (see \ref{subsubsec:Sobel}). Afterwards, we dicuss selection variants of the image's subregions (see \ref{sec:subregion_choice}).

\subsubsection{Edge Intensity Measurement}
\label{subsubsec:Sobel}
Measuring edge intensity is extensively studied in the literature, see e.g., \cite{canny, szeliski2011, dharampal2015, gonzalez2017, Sun2022}. 
The Sobel operator is appropriately balancing efficiency, noise robustness, and accuracy (see \cite{gonzalez2017}), which is justifying its use as the target for DNN-based edge prediction.


It approximates the gradient magnitude of an image using convolution with discrete horizontal and vertical kernels. For a given image \(f \in \mathbb{R}^{3 \times 3}\), the Sobel operator response is computed as:
	\begin{equation}
    \label{eqn:Sobel}
	    G(f) := \sqrt{(G_x *f)^2+(G_y* f)^2},
	\end{equation}
	
	where \(G_x\) and \(G_y\) are the standard Sobel kernels for detecting horizontal and vertical changes in the image`s gray values, respectively. \(G_x\) and \(G_y\) are approximating the first derivative and are defined as:
    
        \[
    G_x := \begin{bmatrix}
    -1 & 0 & +1 \\
    -2 & 0 & +2 \\
    -1 & 0 & +1 \\
    \end{bmatrix}, \quad
    G_y := \begin{bmatrix}
    -1 & -2 & -1 \\
    0 & 0 & 0 \\
    +1 & +2 & +1 \\
    \end{bmatrix}.
    \]

    Furthermore, the convolution \( * \) for an image $f\in \mathbb{R}^{n \times n}$ and a linear kernel $g \in \mathbb{R}^{n \times n}$ is in general defined as (see \cite{szeliski2011}):

    \[
(f * g)[i,j] = \sum_{k} \sum_{l} f[i - k, j - l] \cdot g[k, l]
\]

	This operation yields a scalar value for the center pixel representing the edge intensity within the \(3 \times 3\) image. However, since the Sobel operator formula \eqref{eqn:Sobel} involves square root and quadratic terms, directly integrating it into a MIP would introduce non-convexity. To preserve the tractability of our optimization problem, we train a DNN to approximate the Sobel operator response. Using the MIP formulation (see \ref{subsec:Model}), the trained DNN is then encoded into the optimization problem using binary variables and linear constraints that can be handeled by modern MIP solvers. 

\begin{figure}[htbp]
    
    \centering
    \begin{subfigure}[t]{0.4\textwidth}
        \centering
        \includegraphics[height=4.4cm]{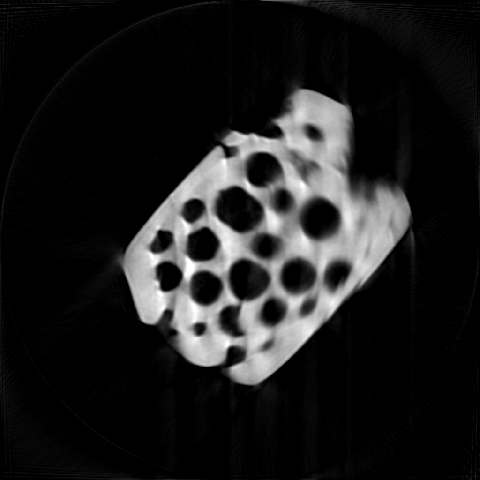}
    \end{subfigure}
    \hspace{0.03\textwidth}
    \begin{subfigure}[t]{0.4\textwidth}
        \centering
        \includegraphics[height=4.4cm]{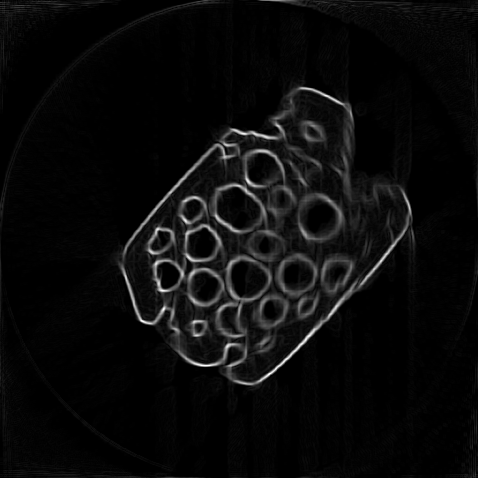}
        
    \end{subfigure}
    \caption{Image of a macroporous zeolite particle (left) vs. gradient image (right) obtained by using the Sobel edge filter given by equation \eqref{eqn:Sobel}.}
    \label{fig:edge_filter_demo}
\end{figure}

\figref{fig:edge_filter_demo} shows a slice through a 3D reconstruction by electron tomography of a macroporous zeolite particle (left) and its edge intensity measured by the Sobel operator (right). The bottom left part of the particle depicts a very binary (black and white) image of the gradient. However, moving to the upper right the original image increasingly smeares and, implicitly, the Sobel operator returns intermediate (gray) values. 

\FloatBarrier

\subsubsection{Subregion Selection}\label{sec:subregion_choice}
We apply the Sobel operator utilizing \(3 \times 3\) kernels via convolution with images of same size and thus partition the image into \(3 \times 3\) pixel subregions. Each subregion introduces new constraints to the MIP model. Accordingly, the frequency with which a pixel is included in multiple subregions directly influences the number of constraints associated with its corresponding variable. Therefore, the extent of which pixels are shared among subregions is a critical factor for the efficiency of our optimization procedure. 
To address this, two spacing strategies are considered and analyzed: an overlapping approach, where subregions share pixels, and a non-overlapping approach, where subregions are mutually exclusive.


\textbf{Non-Overlapping:} Adjacent subregions are completely independent, sharing no pixels.

\textbf{Overlapping :} 
The Sobel response is approximated for each pixel, excluding those located at the image boundaries. 
As a result, each pixel variable may be subjected to up to nine times more constraints due to its inclusion in multiple overlapping subregions.\\


\subsection{Mathematical Model for a Trained DNN}
\label{subsec:Model}

To represent a DNN in a mathematical optimization model, we closely follow the exposition from \cite{fischetti2017deep}. Let an already trained DNN consist of $K + 1$ layers $0, \ldots, K$, where layer 0 corresponds to the input layer and $K$ to the output layer. In particular, this means that we already have a matrix $W$ of weights and a vector $b$ of biases of appropriate dimensions. A layer $k \in \{0, 1, \ldots, K\}$ consists of nodes $1,\ldots,n_k$. We denote by $x_k \in  \R^{n_k}$ the output vector of layer $k$. As usual in DNN, for a layer $k\geq 1$, this output is computed via forward propagation through the network. This means that for each internal node, an activation function is applied to a linear combination of the input of the preceding layer $k-1$, with coefficients given by the corresponding entries in $W,b$. We denote by $W^{k-1}, b^{k-1}$ the weights and biases of the layer $k-1$, respectively. In formulas, the output of layer $k$ is computed as 
\begin{equation}\label{eq:generalformulaNN}
x^k = \sigma (W^{k-1}x^{k-1} + b^{k-1}),\end{equation}
where $\sigma(\cdot)$ denotes the activation function used in the specific layer. We use as activation function the so-called rectified linear unit, ReLU, which for a vector $y$ is computed componentwise via 
$$\text{ReLU} (y) := \max\{0, y\}.$$ 
In order to compute the ReLU output for a vector, we thus 'only' need to decide for each of its entries whether it is smaller than zero or not, as then the output is either zero or the entry itself. In mixed-integer optimization, for each ReLU function, this decision is modeled via a binary variable that either takes value 0 or 1 to differentiate between these two cases. 

For a general equation of the form
$x = \text{ReLU} (w^{\top}y + b)$, 
we first write

\begin{equation}\label{eq:constraintsNN1}
w^{\top} y + b = x - s, \quad x \geq 0, \quad s \geq 0,
\end{equation}
to decouple the positive and negative parts of the ReLU input. In order to avoid that both the positive and the negative parts may become nonnegative, binary $z\in \{0,1\}$ is used: 
\begin{align}\label{eq:constraintsNN2}
z = 1 &\Rightarrow x \geq 0,\notag\\ 
z = 0 &\Rightarrow s\geq 0.
\end{align}

Such implication constraints \eqref{eq:constraintsNN2} can be easily handled by the modern available MIP solvers. We refrain from going into further detail and refer to \cite{fischetti2017deep}. 

We denote by \eqref{eq:DNN} the corresponding mathematical model that consists of linear constraints modeling \eqref{eq:generalformulaNN} for \emph{all} layers $k\in\{0,\ldots, K\}$ via \eqref{eq:constraintsNN1} and \eqref{eq:constraintsNN2}, i.e.:

\begin{equation}\tag{DNN}\label{eq:DNN}
\text{DNN}(x^0)=x^K.
\end{equation}
Hence, the notation $\text{DNN}(\cdot)$ denotes the set of all constraints that are necessary to model a trained neural network using ReLU activation. Therefore, after having trained a neural network, new input $x^0$ can be inserted into $\text{DNN}$ to receive the corresponding output $x^K$.

It is important to note that due to \eqref{eq:constraintsNN2}, the number of activation functions, i.e., number of internal nodes of the neural network determines the number of binary variables in \eqref{eq:DNN}. As the running time to solve a mixed-integer binary problem is largely determined by this number, it is important to balance the size of the neural network with the obtained solution quality and the required running times. We will later discuss this in more detail.

Next, we set up these constraints for a DNN that is trained to learn the Sobel operator. In more detail, we focus on subregions of size $3\times 3$ here, where the generalization to larger subregions is straightforward. 

For some chosen $3\times 3$ (sub-)region, we train a DNN to approximate the Sobel value of the considered subregion. We denote the set of all considered subregions by $A$ and index the subregions by $a\in A$. The nodes' pixel values range between $0$ to characterize void and $\omega$ to characterize material. As mentioned before, a high Sobel value on a subregion $a$ corresponds to an edge, whereas a low Sobel value corresponds to a homogeneous image within the subregion. Hence, if trained appropriately, the same holds for $\text{DNN}$.

Let us denote by $f_a\in [0,\omega]^{3\times 3}$ the matrix containing the pixel values of the subregion, where we recall a pixel value of $\omega$ to denote material. Then,
$$\text{DNN}: [0,\omega]^{3\times 3} \rightarrow \R,\quad  \text{DNN}(f_a)=y_a,$$
maps $f_a$ to an approximate $y_a$ of the Sobel value of the subregion $a$.

Next, for each subregion, we set up an optimization model that aims at improved image reconstruction, by considering constraints given by $\text{DNN}$. To this end, the corresponding objective function aims for reconstructions with either high or low Sobel values, or analogously, either sharp edges or homogeneous material/void. This means that for each subregion it needs to be decided whether it is part of an edge, part of the material or the background. As it is not clear beforehand what values are considered 'high' or 'low', we incorporate a parameter $T$ that controls at what Sobel value the subregion is considered to contain an edge or not. Thus, the objective function aims at deciding this by maximizing the expression $\max\{y_a,T-y_a\}$, where feasible Sobel values $y_a >> T$ indicate an edge in the subregion $a$. In order to linearize this term, we again introduce a binary variable $e_a$ and include the term $e_ay_a + (1-e_a)(T-y_a)$ in the objective function that equals $y_a$ if  $e_a=1$ and $T-y_a$ if $e_a=0$. Where $T$ is a given parameter that will be varied in the computational experiments in Section \ref{sec:compresults}. In formulas, the corresponding optimization model reads as 
\begin{subequations}\label{prob:MIP_single_a}
    \begin{align}
\label{MIP_obj_a}
    \max_{e_a,\,f_a} \quad \text{MIP}_{\text{a}}(e_a, f_a) :=  \max_{e_a,\, f_a}\quad & e_{a} y_{a} + \left(1 - e_{a}\right)(T - y_{a})  \\
       \text{s.t.: } & \text{DNN}( f_a)=y_a ,\label{constr:NN_single_subregion}\\
       & e_{a} \in \{0, 1\},\ y_a \in \R_{\geq 0},\\
       & f_a \in [0,\omega]^{3\times 3}.\label{constr:MIP_single_a_boundedness}
    \end{align}
\end{subequations}
It is worth noting that without further constraints or objectives on $f_a$ like the ones illustrated in \eqref{prob:cshm}, the model \eqref{prob:MIP_single_a} is solely going to construct a $3\times 3$ image of an edge if $T$ is low by setting $e_a=1$ and a homogeneous $3\times 3$ image of black or white color otherwise by setting $e_a=0$. How to balance \eqref{prob:MIP_single_a} with other objectives is going to be addressed in the upcoming Sections \ref{sec:framework_integrated} and \ref{sec:framework_reopt}. Furthermore, we like to highlight that even though the DNN is defined by several other internal variables like \( x^k_{a,j},s^k_{a,j}, z^k_{a,j}\), all these variables are indirectly defined by \(f_a\). Consequently, the MIP \eqref{prob:MIP_single_a} only needs to be optimized for \(f_a\) and \(e_a\), which eases the presentation significantly, but has to be kept in mind, when assessing numerical complexity.

The extension of \eqref{prob:MIP_single_a} to a larger set of subregions $A$, e.g., given by the overlapping approach from Section \ref{subsec:NN_Design} is now rather straightforward. We extend the binary decisions to a vector $e\in \{0,1\}^{A}$, the Sobel values to a continuous vector $y\in \R_{\geq 0}^{A}$, the pixels to a vector $f \in [0, \omega]^{A\times 3\times 3}$ and add $|A|$ many constraints of form \eqref{constr:NN_single_subregion} for each subregion. The objective function then is extended to sum up $|A|$ many terms of form \eqref{MIP_obj_a}:

\begin{subequations}
\label{prob:MIP_edge_enhancement}
    \begin{align}
\label{MIP_obj_func_SUM_MIP_a}
    \max_{e,\,f} \quad \text{MIP}_{\text{A}}(e, f) :=  \max_{e,\, f}\quad & \sum_{a\in A}\left[ e_{a} y_{a} + \left(1 - e_{a}\right)(T - y_{a}) \right]  \\
       \text{s.t.: } & \text{DNN}( f_a)=y_a \quad \forall a\in A ,\label{constr:NN_multiple_subregions}\\
       & e \in \{0, 1\}^{A},\ y \in \R_{\geq 0}^{A},\\
       & f \in [0,\omega]^{A\times 3\times 3}.\label{constr:MIP_edge_enhancement_3}
    \end{align}
\end{subequations}
We note in passing that, at this point, \eqref{prob:MIP_edge_enhancement} may be easy enough so that an optimum solution can be read off easily. However, since \eqref{prob:MIP_edge_enhancement} is only a subproblem for the upcoming reconstruction models, these read-off solutions will be non-optimal for the final model. In addition, it is worth noting that the computational effort required to solve such a MIP grows significantly with the number and size of the considered subregions as well as the number of neurons used within the DNN to approximate the edge intensities. Hence, one may only be able to incorporate a subset of constraints of type \eqref{constr:NN_multiple_subregions} into the DNN-MIP reconstruction model, and thereby enforce an enhanced edge intensity in specific subregions.

\subsection{Model I: DNN-MIP Integration}\label{sec:framework_integrated}
The first model we consider is to integrate the MIP \eqref{prob:MIP_edge_enhancement} into the CSHM model \eqref{prob:cshm}. To this end, it is important to note that \eqref{prob:MIP_edge_enhancement} considers the variable $f$ to be scaled between $0$ and $\omega$, whereas \eqref{prob:cshm} has the freedom to choose pixel values that fit the raw data best, even if this is achieved by certain pixel values exceeding $\omega$. Note that the input for the constraints in \eqref{prob:MIP_edge_enhancement} must adhere to the requirement that its value falls between $0$ and $\omega$ as this is the specific range within which our DNN model has been trained to make predictions; inputs outside this range are likely to be predicted poorly. Hence, in order to synchronize the two approaches, we therefore first run CSHM indepently to determine the maximal pixel intensity of CSHM
$$f_{\text{max}} \coloneqq \max_{j\in I}\ \tilde{f}_j.$$

Subsequently, given a constant weight factor $\phi \geq 0$, we integrate \eqref{prob:MIP_edge_enhancement} into CSHM \eqref{prob:cshm} as follows:
    \begin{subequations} 
        \label{CSHM_MIP_integrated}
        \begin{align}
            \min_{e , \tilde{f}} \quad \operatorname{CSHM-MIP}(e,f) := \min_{e,\tilde{f},f}\ & \text{CSHM}(\tilde{f}) - \phi\text{MIP}_A(e,f), \label{f_tilde_normalized_CSHM} \\
            \text{s.t.: } & f  =  \frac{\omega \tilde{f}}{f_{\text{max}}},  \\
            & \eqref{constr:CSHM1}-\eqref{constr:CSHM4},\\
            & \eqref{constr:NN_multiple_subregions} -\eqref{constr:MIP_edge_enhancement_3}.
        \end{align}
    \end{subequations}
Here, \(\phi\) is a weighting factor that balances the influence of the DNN-MIP objective relative to the CSHM objective.
Due to the penalizations in the objective function, the model balances between favoring solutions close to the CSHM values and penalizing deviations from the learned Sobel predictions. In summary, given the maximal pixel value $f_{max}$ and the parameters $\phi, T$, we solve \eqref{CSHM_MIP_integrated} in order to determine an edge-enhanced image compared to the CSHM solution given by \eqref{prob:cshm}. Moreover, the integrated model \eqref{CSHM_MIP_integrated} demonstrates how the above DNN-MIP constraints and objectives can be integrated into an existing reconstruction algorithm. The disadvantage of the integrated model may be long runtime for modern MIP solvers, depending on image resolution (i.e. number of pixels), considered subregions and number of neurons required in the DNN. This is illustrated by an example in Section \ref{sec:Comparison_integrated_MIPRO}. 
\subsection{Model II: DNN-MIP Re-Optimization}\label{sec:framework_reopt}
The approach \emph{DNN-MIP Re-Optimize Model (MIP RO)} outlined here offers a significantly reduced computational effort compared to \eqref{CSHM_MIP_integrated}. As we consider post processing, we first solve CSHM \eqref{prob:cshm} to determine its optimal solution $f^*$ as well as $f_{max}$. The overall idea of MIP RO is to find an edge enhanced solution, that is on the one hand, close to $f^*$, but on the other hand, mimics the binary structure of the observed material. To this end, we consider three different objective terms and, in order to avoid the computational burden that has to be considered for large images, we consider $3\times 3$ subregions separately. 

First, in order to provide enhanced edges, we consider $\text{MIP}_{\text{a}}$ from 
\eqref{prob:MIP_single_a}. 
Second, we incorporate information about the material density, similar to Constraint \eqref{constr:CSHM1}. 
To this end, we minimize the term
\begin{equation}
\label{density deviation minimization}
\operatorname{Dev}(f_a):= |\omega - f_a||f_a|,
\end{equation}
since its minimization enforces \( f \) to take values close to either \( \omega \) or $0$.
The absolute values from the 1-norm can be dropped since $0\leq f_a \leq \omega$ due to \eqref{constr:MIP_single_a_boundedness}.

Third, to ensure that the optimized image remains close to the original CSHM reconstruction $f^*$, we introduce a pixel loss function as follows:
\begin{equation}
\label{pixel loss function}
\operatorname{L}(f_a):=  \left\|f_a-\frac{\omega f^*_a}{f_{max}}\right\|_2^2,
\end{equation}
where $f^*_a$ denotes the entries of $f^*$ that correspond to subregion $a$. We note that similarly to \eqref{CSHM_MIP_integrated}, we have rescaled $f^*$ to the domain $[0,\omega]^{3\times 3}$ of $f_a$. Moreover, as we minimize the above loss, this adds a convex term to our joint objective function.

We will now incorporate the terms from \eqref{density deviation minimization} and \eqref{pixel loss function} into the objective of \eqref{prob:MIP_single_a}. This leads to the following MIP formulation, which reoptimizes a subvector $f^*_a$ of the CSHM solution $f^*$:

\begin{subequations}
 \label{MIP RO objective function for subregion a}
    \begin{align}
        \max_{e_a,\, f_a} \text{MIP}_{RO}(e_a, f_a) := \max_{e_a,\, f_a}\ & \text{MIP}_{a}(e_a, f_a) - \alpha\operatorname{Dev}_d(f_a) - \beta \operatorname{L}_p(f_a),\\
        \text{s.t.: } & \eqref{constr:NN_single_subregion} -\eqref{constr:MIP_single_a_boundedness},
    \end{align}
    \label{eq:MIP-RO}
\end{subequations}
with weightings $\alpha,\beta \geq 0$ of the corresponding objective functions. The goal is to enhance potential edges of the specimen while preserving image fidelity. 

Next, instead of solving all the subregions simultaneously, 
we optimize each subregion independently and subsequently merge them into a single image. 
To this end, we propose the following algorithm, whose subregions are chosen based on the approaches illustrated in Section \ref{sec:subregion_choice}:

\begin{algorithm}[H]
\caption{Sliding Window MIP Re-Optimization}
\begin{algorithmic}[1]
\Require Image $f^* \in \R^{M \times N}$, subregion spacing $s \in \{1,3\}$, MIP config $(\alpha, \beta, T)$, trained neural network for subregions $\text{DNN}$
\Ensure Optimize CSHM image $f^*$ into $\hat{f}$
\State Initialize $\hat{f} \gets $ $M\times N$ array of empty lists
\For{$x \in \{0, s, 2s, \ldots, M-2\}$}
    \For{$y \in \{0, s, 2s, \ldots, N-2\}$}
        \State $a \gets [x:x+2, y:y+2]$
        \State Solve $\text{MIP}_{RO}(e_a,f_a)$ with $f^*_a$, DNN, $T, \alpha, \beta,$ according to \eqref{MIP RO objective function for subregion a}
        \State $(e_a',f_a') \gets \text{argmax}\text{ MIP}_{RO}$  
        \For{$(i,j)$ in $\{0,1,2\} \times \{0,1,2\}$}
            \State $\hat{f}[x+i, y+j] \gets \hat{f}[x+i, y+j] \cup f_a'[x+i, y+j]$
        \EndFor
    \EndFor
\EndFor
\For{each $(i,j)$ in $M\times N$ }
    \State $\hat{f}[i, j] \gets $ average of stored values in $\hat{f}[i, j])$
\EndFor
\State \Return{$\hat{f}$}
\end{algorithmic}
\label{alg:sliding_window}
\end{algorithm}

If \eqref{eq:MIP-RO} is solved for overlapping subregions $a$, i.e., for $s=1$, it may reconstruct different pixel values for the pixels that are contained in the overlap of two different subregions. Algorithm \ref{alg:sliding_window} resolves this conflict by assigning the average of all potential pixel values computed by \eqref{eq:MIP-RO}. We note that there are possibilities to finetune Algorithm \ref{alg:sliding_window} to potential user's needs, e.g., one may want to choose the maximal pixel value if one suspects that the raw data may underestimate the amount of material present in the sample. 
Moreover, we would like to mention that line 5 could be executed in parallel in order to further reduce computing time significantly. However, since the article's focus is to assess reconstruction quality rather than efficient execution, we did not execute upon parallelization for the computational results in Section \ref{sec:compresults}.  Since Algorithm \ref{alg:sliding_window} requires a reconstructed CSHM image, the overall runtime can be expected to be at least the runtime of CSHM. However, since a single execution of line 5 was observed to run within a few seconds, it can be expected that the runtime of Algorithm \ref{alg:sliding_window} can be reduced to a slightly slower but competitive runtime compared to CSHM given that a sufficient number of cores is available.


\subsection{Parameter Selection} \label{subsec:parameter}
At the beginning, we would like to state that 
the DNN used here has an input layer containing 9 nodes, two hidden layers with 9 nodes each, and an output layer with a single node. The objective of the MIP RO model \eqref{MIP RO objective function for subregion a} contains three parameters that need to be carefully chosen, and thus, are discussed in the following. 

\subsubsection{Threshold Value \textit{T}}
\label{subsubsec:threshold}
For a given subregion $a \in A$, the threshold value $T$ influences the objective function \eqref{MIP_obj_a} via the term 
\[\max(y_a, T-y_a).\]
This means that the optimizer decides that there is no edge if the subregion`s pixel values $f_a$ result in an approximated Sobel value $DNN(f_a) = y_a < T$ and decides for an edge, if it is sufficiently sharp, i.e. if $y_a > T$.
In order to balance the contribution of both of these effects on the corresponding objective function, we search for a viable $T$ with 
\[T < 2\bar{u},\] 
where \(\bar{u}:= \max_{x\in [0,\omega]^9} DNN(x)\) as this is the maximal Sobel value attainable by DNN. 
In our computational results, we compute $\bar{u}=550$ by solving $\max_{x\in [0,\omega]^9} DNN(x)$ by Gurobi, thereby reducing our search space to $T<1100$. After an exhaustive hyperparameter study, $T=800$ arose as an effective choice for the threshold allowing for minor fluctuations in the material while detecting and highlighting edges effectively.

\subsubsection{Choosing Values for \textit{$\alpha,\beta$}}
\label{subsubsec:alpha,beta}
The DNN-MIP term \eqref{MIP_obj_a} contributes linearly to the objective and involves binary decision variables, while the material property deviation term~\eqref{density deviation minimization} introduces non-convex quadratic and the pixel loss term~\eqref{pixel loss function} convex quadratic dependencies. We are flexible on how to choose the relation between the weighting parameters $\alpha$ and $\beta$. 
When they are chosen to be equal, i.e., $\alpha = \beta$, and assuming a CSHM reconstruction \(f^\star\), the combined contribution of the material property and pixel loss terms simplifies to a linear expression in
\(f\). Solving MIPs with linear objectives is considerably faster than when they are non-linear. Indeed, setting $\alpha=\beta=1$, we have:
\begin{subequations}
\begin{align*}
\label{sum of Dev and LP linear proof}
\alpha \cdot \operatorname{Dev}_d(f) + \beta \cdot \operatorname{L}_p(f) &= \quad \operatorname{Dev}_d(f) + \operatorname{L}_p(f) \\
&=  \quad (\bar{u} - f)f +  \|f-f^\star\|_2^2 \\
&= \quad \bar{u} \cdot f - f^2 + f^2 + (f^\star)^2 - 2 \cdot f \cdot f^\star \\
&= \quad (\bar{u} - 2 \cdot f^\star)  \cdot f + (f^\star)^2,
\end{align*}    
\end{subequations}
which is affine in \(f\),  since  \(\|f^\star\|_2^2\) is a constant.
We summarize this
as follows:
\begin{enumerate}
\item \textbf{Case \(\alpha = \beta\)}:
The overall objective function is linear with respect to \(f\), and the MIP RO reduces to a binary linear program. It can be solved 
to global optimality even for huge instances 
using standard MIP solvers. 

\item \textbf{Case \(\alpha < \beta\)}:  
The convex quadratic pixel loss term dominates. Since it is convex with respect to \(f\), the model in principle can still be solved by MIP solvers, but requires longer running times. 

\item \textbf{Case \(\alpha > \beta\)}:  
The problem is non-convex, as the non-convex quadratic material deviation term dominates. This potentially results in large running times for modern MIP solvers.
\end{enumerate}

\FloatBarrier

\section{Experimental Details}
\label{sec:exp-details}

\subsection{Reconstruction Benchmark and Hardware}\label{sec:Hardware}
Results are compared against four established image reconstruction algorithms: 
SIRT with non-negativity constraint, CS, TVR-DART, and 
CSHM. CS refers to the solution of \eqref{TVRdartmodel} using the same solver - Gurobi Optimizer Version 11.0, see \cite{gurobi} - and the same parameter values as in CSHM, but without CSHM's additional constraints described in \cite{Kreuz}. 
Gurobi is run with a solution tolerance of $10^{-6}$. For the comparison algorithms, the number of reconstruction iterations is set to 1000 for SIRT and 250 for TVR-DART, as used in \cite{Kreuz}. The initial SIRT reconstruction used as input for TVR-DART is also computed with 1000 iterations. Note that TVR-DART may terminate earlier if a convergence stopping criterion is satisfied before reaching the maximum number of iterations. For the integrated approach, we used $\phi = 1 \cdot 10^8$, 
although this value depends on the resolution, the dataset, and the specific algorithm with which the MIP equations are integrated.

All reconstruction experiments were carried out on a local workstation equipped with an 8-core (16-thread) CPU, 32 GB of RAM, and an NVIDIA GeForce RTX 3080 GPU with 8 GB VRAM, running Windows 11. Python 3.7.12 was used as the primary software environment supporting implementation and benchmarking of all algorithms (SIRT, TVR-DART, CS, CSHM and MIP RO). Gurobi Optimizer version 11.0, see \cite{gurobi} served as the solver of the resulting problem instances. It is worth noting that Gurobi has recently added its Gurobi Machine Learning package that can be used to create constraints of type \eqref{constr:NN_multiple_subregions}, which could be utilized to improve computational performance. However, since this article focuses on a thorough investigation of the underlying methodology and image quality rather than optimal computational performance, we chose not to use the built-in package.


Additional batch-scale reconstructions (see Supplementary Video 3), and parallel simulations were executed on FAU's high-performance computing cluster on 2 Intel Xeon Gold 6326 processors ("Ice Lake", 16 cores @2.9 GHz) and 256 GB RAM, utilizing multiple nodes with CPU-GPU hybrid architectures to accelerate large-scale optimization tasks.
    
\subsection{Tomographic Datasets}   
\begin{figure}[htbp]      
    \centering
     \begin{subfigure}[b]{0.3\textwidth}
            \centering
            \includegraphics[width=\textwidth]{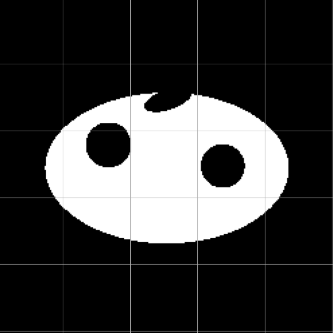}
            \caption{Simulated phantom}
            \label{dataset-a}  
        \end{subfigure}
        \begin{subfigure}[b]{0.3\textwidth}
            \centering
            \includegraphics[width=\textwidth]{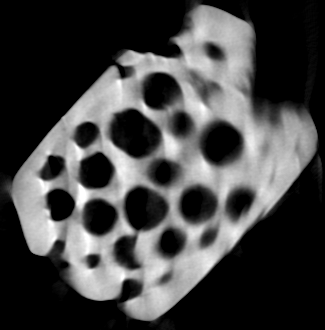}
            \caption{Porous zeolite particle}
            \label{dataset-b}  
        \end{subfigure}
        \begin{subfigure}[b]{0.3\textwidth}
            \centering
            \includegraphics[width=\textwidth]{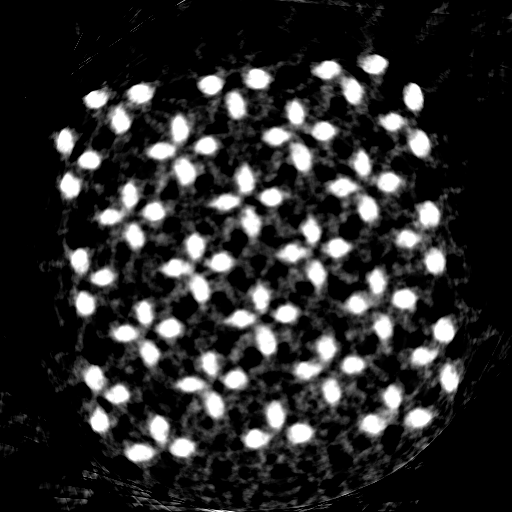}
            \caption{Cu microlattice}
            \label{dataset-c}  
        \end{subfigure}
    \caption{Three different datasets used in this study: (a) simulated phantom structure (ground
truth used for relative mean error (RME) calculation), (b) electron tomography reconstruction of a porous zeolite particle (SIRT with 1000 iterations using 180 projections in a 180° tilt-angle range), (c) absorption contrast nano-CT reconstruction of copper microlattice structure (SIRT with 1000 iterations using 40 projections in a 180° tilt-angle range).}
    \label{dataset}  
\end{figure}

    The image reconstructions discussed in this article were carried out using three datasets  shown in \figref{dataset}: a simulated phantom object slice including random Poisson noise, along with two experimental datasets — one obtained via ET and the other through nano-CT. These datasets were also used in \cite{Kreuz}.

   The simulated phantom dataset shown in \figref{dataset-a} consists of a two-dimensional (2D) elliptical structure with two circular holes, representing a homogeneous material with a single constant density value, $\omega$, within the material region. It was used to evaluate reconstruction algorithms at a pixel number of $64^2$ and $512^2$ pixels under two sparse acquisition scenarios using 5 and 20 projection angles, which were equidistantly selected from a full set of 180° tilt-angle range. This setup allows us to simulate data sparsity, mimicking cases where full angular coverage is not available while still providing us a ground truth for quantitative comparison. To simulate realistic imaging conditions, we added Poisson noise -- often referred to as shot noise \cite{kak2001principles,lee2018} -- to the complete set of projections prior to subsampling. Since the ground truth of the simulated phantom is noise-free, this introduction of noise captures the inherent randomness in photon or electron detection, enabling a more realistic comparison with experimentally acquired data.
   
\begin{figure}
    \centering
    \includegraphics[width=1\linewidth]{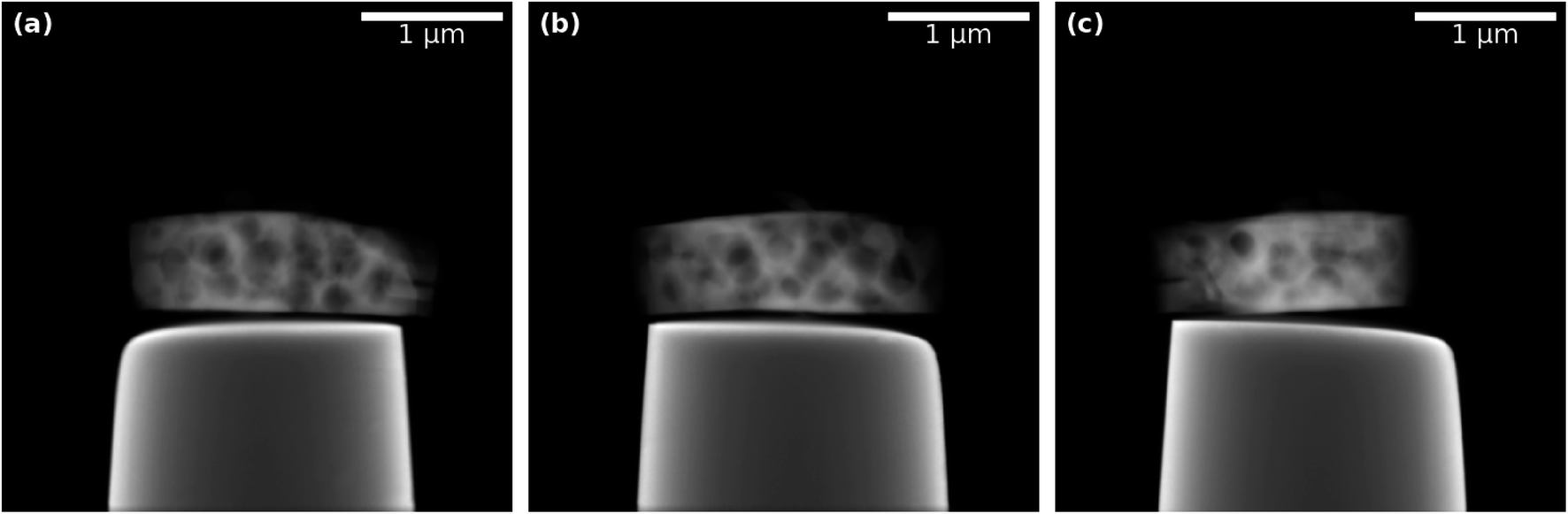}
    \caption{Exemplary projections of experimental ET data from different projection angles of a macroporous zeolite particle on the plateau of a tomography tip: (a) 20°, (b) 90°, (c) 160°.}
    \label{proj-showing-et1}
\end{figure}

The ET data shown in \figref{dataset-b} and \figref{proj-showing-et1} consists of a macroporous MFI-type zeolite particle synthesized by \cite{machoke2015micro}. A complete tilt series was collected, as demonstrated in Supplementary Video 1, covering a tilt-angle range of 180° with 1° increments (resulting in a total of 180 projections). This was performed on a particle positioned at the top of a tomography tip (referred to as 360°-ET or on-axis ET) using a FEI Titan$^{3}$ 80-300 transmission electron microscope, operated at an acceleration voltage of 200 kV in high-angle annular dark-field (HAADF) scanning transmission electron microscopy (STEM) imaging mode. The images captured were sized at 1024 x 1024 pixels, with a pixel size of 3.55 nm. A Fischione Model 2050 On-Axis Rotation Tomography sample holder (E.A. Fischione Instruments, Inc.) was utilized for the tomography. The HAADF-STEM imaging technique ensures approximately parallel beam geometry and establishes a direct relationship between measured intensities, sample mass (density), and thickness, allowing for the application of Radon transform in the reconstruction process. For additional experimental details, please refer to \cite{przybilla2018transfer}.

\begin{figure}
    \centering
    \includegraphics[width=1\linewidth]{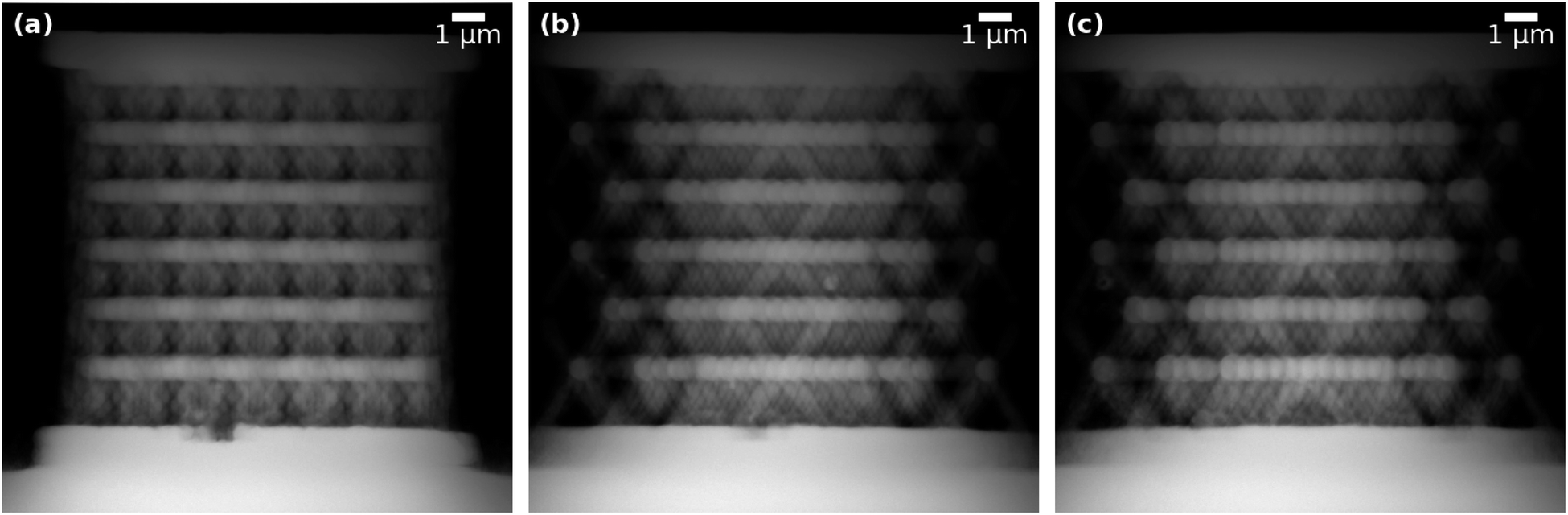}
    \caption{Exemplary projections of experimental absorption-contrast nano-CT data set from different projection angles of a copper microlattice structure: (a) 0°, (b) 10°, (c) 30°.}
    \label{proj-showing-et2}
\end{figure}

The nano-CT dataset illustrated in \figref{dataset-c} and \figref{proj-showing-et2} features a copper microlattice. This microlattice was produced using an additive micromanufacturing technique that employs localized electrodeposition in a liquid medium (CERES system – Exaddon AG, Switzerland) and are described in more detail in \cite{kang2023fabrication, ramachandramoorthy2022anomalous,Kreuz}. 
The nano-CT experiment was conducted using a ZEISS Xradia 810 Ultra X-ray microscope, which features a 5.4 keV rotating anode chromium source operating in absorption contrast and large-field-of-view (LFOV) mode. In LFOV mode, the microscope achieves a field of view (FOV) of 65 µm x 65 µm, with spatial resolution down to 150 nm and a pixel size of 63.89 nm. A tilt series comprising 40 projections was collected over a tilt-angle range of 180° (with a tilt increment of 4.5°), employing an exposure time of 300 seconds per frame, as shown in Supplementary Video 2. Assuming parallel beam geometry and beam attenuation described by the Beer-Lambert law, the Radon transform was applied for reconstruction using the projection data represented as $p = - \ln{I/I_{0}}$, where $I$ denotes the measured intensity and $I_{0}$ represents the unattenuated incident beam intensity. The nano-CT projection images displayed in \figref{proj-showing-et2} are shown with inverted contrast.  \\
The 360°-ET tilt series was aligned using cross-correlation in FEI Inspect 3D version 3.0. In contrast, the nano-CT tilt series underwent alignment through the adaptive motion compensation 
procedure, as described by  \cite{Wang2012}, which was implemented in the native ZEISS software (XMController). \\
For image reconstruction, the projections of all three datasets were selected evenly distributed across the available angular range. Specifically, they range from 0° to 179° when there is no missing wedge. In the case of a missing wedge of $x$°, the angles are distributed from $\frac{x}{2}$° to $(180-\frac{x}{2})$°.

\subsection{DNN Training Data}

\begin{figure}[htbp]    
    \centering
    \begin{subfigure}[t]{0.45\textwidth}
        \centering
        \includegraphics[width=\textwidth]{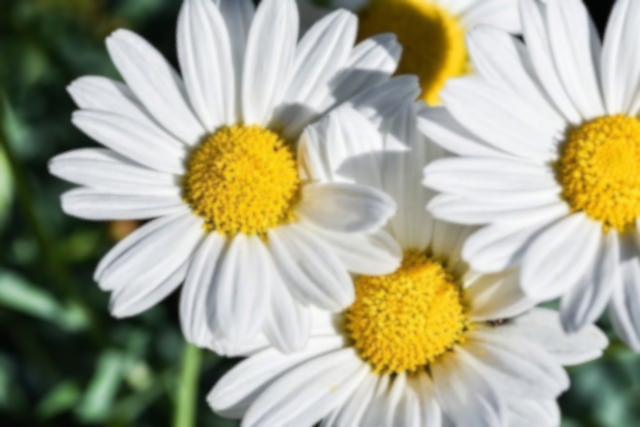}
        \caption{\footnotesize Original Flower Image (training data for input vector)}
        \label{fig:flower}
    \end{subfigure}
    \begin{subfigure}[t]{0.45\textwidth}
        \centering
        \includegraphics[width=\textwidth]{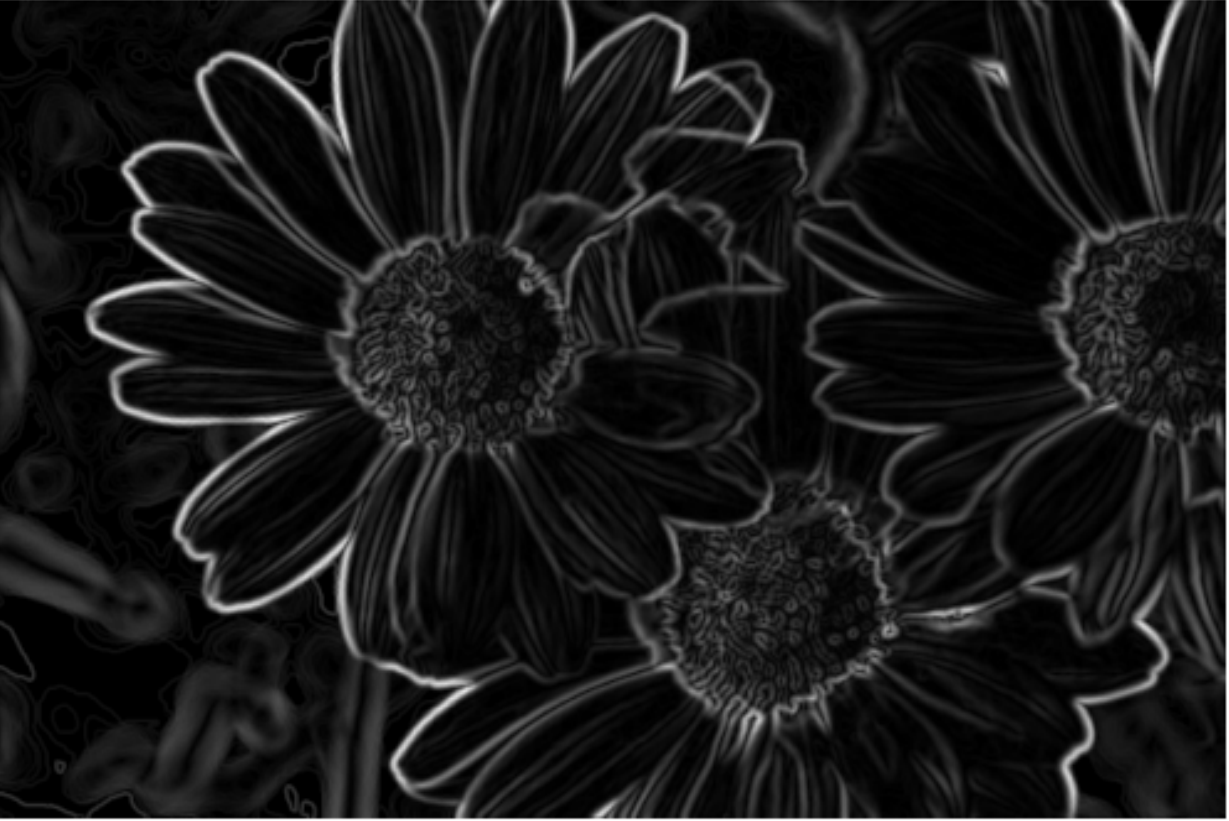}
        \caption{\footnotesize Sobel Edge Map of Flower Image (training data for output vector)}
        \label{fig:flower edge}
    \end{subfigure}
    \caption{\footnotesize Edge detection results using the trained DNN. Training input: normalized grayscale version of (a), training output: Sobel operator output of (a), see (b) for an illustration. (image source for (a): https://pixabay.com/photos/daisies-nature-bloom-white-summer-9637979/)
    }
\end{figure}

To train the DNN, a grayscale image of flowers was first converted into a suitable format by dividing it into \(3 \times 3\) subregions, with each pixel serving as the center of a subregion. The model was trained to predict the edge intensity of this subregion, using Sobel operator values as reference. 
This image was chosen for training because it contains a diverse range of edge orientations and intensity variations. The corresponding Sobel values were normalized to ensure a compact activation value range. After training, the DNN's edge detection performance was evaluated on a variety of other grayscale images to confirm its effectiveness in detecting edges across different visual contexts. The considered DNN is provided in the supplementary material.

\section{Computational Results}\label{sec:compresults}
\subsection{Reconstruction Quality Descriptors}
To assess reconstruction quality objectively, we computed the relative mean error (RME) between the pixel values \( f \) of the reconstructed image and the corresponding pixel values \( \hat{f} \) in the original image, i.e. the ground truth of the phantom (\figref{dataset-a}). The RME is defined as

\begin{equation}
\text{RME}(f) := \frac{\sum_{j=1}^{n} \left| {f_j} - \hat{f}_j \right|}{\sum_{j=1}^{n} \left| \hat{f}_j \right|}.
\end{equation}

Whereas in case of the experimental datasets with no ground truth, we use raw data coverage (RDC) as a measure on how well the reconstruction aligns with the experimental projection dataset, where \( R \in \mathbb{R}^{m \times n} \) is the discretized Radon transform matrix, \( f \in  \mathbb{R}^ n \) are the pixel values in the reconstructed image and \( \hat{p}\in  \mathbb{R}^m \) is the measured projection data. The RDC is given by

\begin{equation}
\text{RDC}(f) := \frac{\sum_{i=1}^{m} \left| {(Rf)}_i - \hat{p}_i \right|}{\sum_{i=1}^{m} \left| \hat{p}_i \right|}.
\end{equation}

As an indicator for the strength of binarization and contrast in the reconstruction, we introduce the bimodal contrast score (BMS) \cite{otsuBMS}; \cite{gonzalez2017}, which quantifies the proportion of pixel intensities that lie near the extreme values (0 and $f_{max} := \max f_i$):
\begin{equation}
\text{BMS}(f) := \frac{1}{n} \sum_{i=1}^{n} \left[ \mathbbm{1}_{[0,\epsilon]}(f_i) + \mathbbm{1}_{[f_{max}-\epsilon, f_{max}]}(f_i) \right]
\end{equation}
where $\mathbbm{1}_{[a,b]}(\cdot)$ is the indicator function on an interval $[a,b] \subset \mathbb{R}$, i.e.,
$$\mathbbm{1}_{[a,b]}(f)=\begin{cases}
    1 & \text{ if } f\in [a,b]\\
    0 & \text{ otherwise.}
\end{cases}$$ 
In this study, the threshold \( \epsilon = 10 \) was empirically chosen to capture pixel intensities sufficiently close to the extremes (0 and 255), balancing sensitivity to high-contrast regions without being overly sensitive to minor noise.
 A BMS of 1 indicates an almost completely binarized image, whereas a low BMS close to 0 resembles an image containing a broad variety of greyscale values. \\

Furthermore, to illustrate and quantify the impact of the selection NN threshold value $T$ from objective \eqref{MIP_obj_a} on the final reconstructed image (see \figref{TinfluenceArea}), we calculate the material coverage (MC), i.e. the number of non-zero pixel values in the reconstruction: 

\begin{equation}
\text{MC}(f) := \sum_{i=1}^{n}\mathbbm{1}_{(0,f_{max}]}(f_i). 
\end{equation}

The higher the value of MC, the smaller the pore space and more pixels assigned to a material are present in the reconstruction. \\
Note that neither RDC, BMS or MC are as reliable as RME for reconstruction quality comparison.

\subsection{Comparison of CSHM, Integrated Approach and MIP RO Approach} \label{sec:Comparison_integrated_MIPRO}

\begin{figure}[htbp]
\small
    \centering
    \begin{subfigure}[t]
    {0.24\textwidth}
        \centering
        \includegraphics[width=\textwidth]{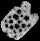}
        \caption{CSHM: 3 s, BMS: $0.795$}
        \label{fig:firstcomparison-a}
    \end{subfigure}
    \begin{subfigure}[t]{0.24\textwidth}
        \centering
        \includegraphics[width=\textwidth]{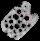}
        \caption{Integrated: 600 s,\\ $\phi=1e8$, BMS: $0.797$}
        \label{fig:firstcomparison-b}
    \end{subfigure}
    \begin{subfigure}[t]{0.24\textwidth}
        \centering
        \includegraphics[width=\textwidth]{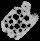}
        \caption{MIP RO: 63 s,\\ $\alpha=\frac{1}{125}$, BMS: $0.872$ }
        \label{fig:firstcomparison-c}
    \end{subfigure}   
    \begin{subfigure}[t]{0.24\textwidth}
        \centering
        \includegraphics[width=\textwidth]{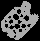}
        \caption{MIP RO: 9 s,\\ $\alpha=\frac{1}{50}$, BMS: $0.990$ }
        \label{fig:firstcomparison-d}
    \end{subfigure} 
    \caption{ \small Comparison of CSHM, Integrated Framework and MIP RO with $64^2$ pixels and 20 projections, $T=800$, and $\beta=\frac{1}{50}$. The integrated approach was stopped after 600 s at $15$\% optimality gap with $T=200$.}
    \label{fig:firstcomparison}
\end{figure}

\figref{fig:firstcomparison} shows a comparison of the same slice through the zeolite particle from the electron tomography tilt series shown in \figref{proj-showing-et1} reconstructed by CSHM, the integrated approach, and two MIP RO instances at a pixel number of $64 \times 64$. The respective runtimes and BMS are indicated.

The CSHM reconstruction in \figref{fig:firstcomparison-a} serves as a benchmark to compare the two new approaches. Moreover, it serves as reference for MIP RO shown in \figref{fig:firstcomparison-c} and \figref{fig:firstcomparison-d}. \figref{fig:firstcomparison-b} displays the results of the integrated approach. Due to its high runtime, it is only applied to the region of interest (ROI) indicated by the transparent red rectangle. The integrated approach delivers a significantly clearer and sharper pore compared to the same ROI of the CSHM reconstruction. The MIP RO reconstructions in \figref{fig:firstcomparison-c} and \figref{fig:firstcomparison-d} demonstrate that the degree of binarization in the image can be influenced by tuning parameter $\alpha$ and $\beta$ from equation \eqref{eq:MIP-RO} -- a high degree of binarization means that pores and zeolite phase both mostly exhibit constant intensity values, leading to a BMS of almost 1. This can be achieved by choosing $\alpha = \beta$. As already explained in \ref{sec:framework_integrated}, the integrated model is computationally expensive and therefore not suited for the reconstruction of larger images or images at higher resolution. However, it leads to convincing results and might be valuable for enhancement of limited regions of the overall picture.
Nevertheless, in the following, we focus solely on how the MIP RO approach compares to CSHM, and SIRT and TVR-DART as established reconstruction algorithms. 

\FloatBarrier

\subsection{Results for Simulated Phantom} 

\begin{figure}[htbp]
        \scriptsize
        \begin{subfigure}[t]{0.19\textwidth}
            \centering
            \includegraphics[width=\textwidth]{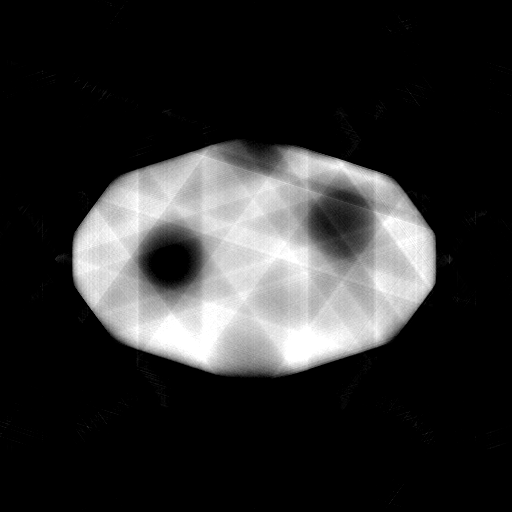}
            \caption{SIRT: 15 s, \\ RME: $0.264$, BMS: $0.764$}
            \label{phantom_5proj-a}
        \end{subfigure}
        \begin{subfigure}[t]{0.19\textwidth}
            \centering
            \includegraphics[width=\textwidth]{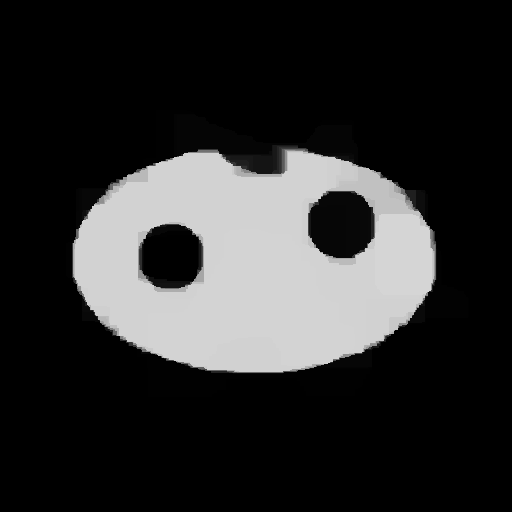}
            \caption{CS: 176 s,\\RME: $0.053$, BMS: $0.963$}
            \label{phantom_5proj-b}
        \end{subfigure}
        \begin{subfigure}[t]{0.19\textwidth}
            \centering
            \includegraphics[width=\textwidth]{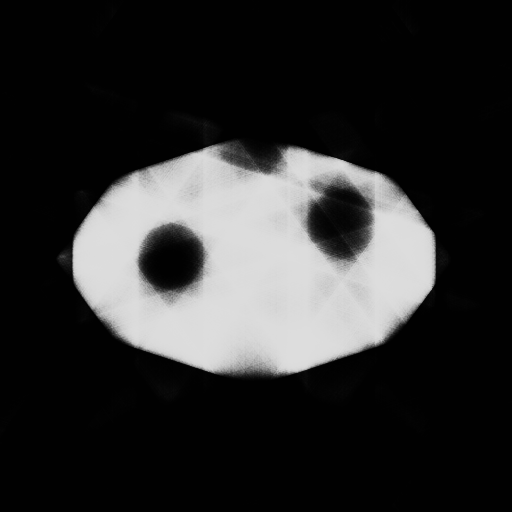}
            \caption{TVR-DART: 29 s,\\  RME: $0.105$, BMS: $0.895$}
            \label{phantom_5proj-c}
        \end{subfigure}
        \begin{subfigure}[t]{0.19\textwidth}
            \centering
            \includegraphics[width=\textwidth]{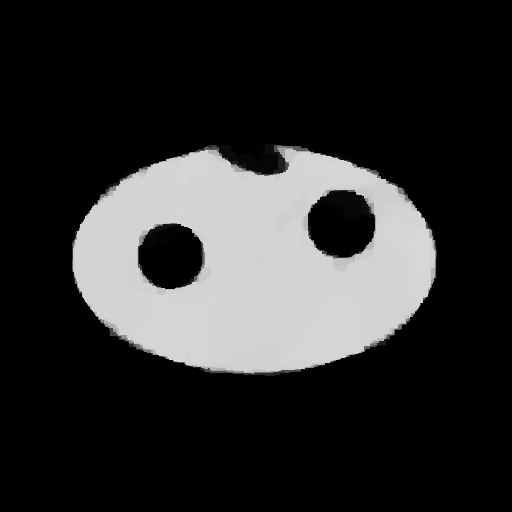}
            \caption{CSHM: 84 s,\\ RME: $0.042$, BMS: $0.974$ }
            \label{phantom_5proj-d}
        \end{subfigure}
        \begin{subfigure}[t]{0.19\textwidth}
            \centering
            \includegraphics[width=\textwidth]{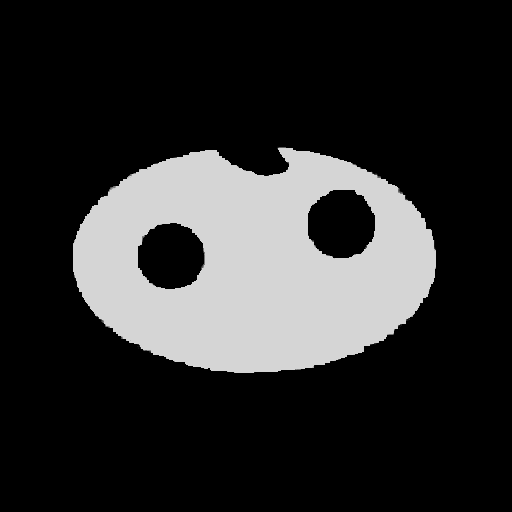}
            \caption{MIP RO: 1388 s,\\ RME: $0.017$, BMS: $0.998$ }
            \label{phantom_5proj-e}
        \end{subfigure}
    \caption{\small Results for the simulated data with $512^2$ pixels and 5 projections -- $\lambda_{cs}=20000$, $\lambda_{TVR}=500$, $T=800$, $\alpha=\frac{1}{50}, \beta=\frac{1}{50}$.}   
    \label{phantom_5proj}
\end{figure}

First, we compare reconstructions of the simulated data set including random Poisson noise with $512^2$ pixels and from only $5$ projection angles in a tilt-angle range of 180° with equidistant tilt increment (i.e., no missing wedge). The results are displayed in \figref{phantom_5proj}, where corresponding runtimes, RMEs and BMS values are noted below the reconstructions, alongside with the applied
reconstruction parameters of $T$, $\alpha$, $\beta$ (MIP RO), $\lambda_{cs}$ and $\lambda_{TVR}$.

Despite the simplicity of the object's shape and the presence of only random Poisson noise in the projections -- without any additional inconsistencies, such as tilt series misalignment or nonlinear contrast variations -- the limited number of projections is insufficient for a satisfactory SIRT reconstruction. As shown in \figref{phantom_5proj-a}, the resulting image suffers from significant streaking artifacts and non-homogeneous intensities, albeit with a quick runtime. This is also reflected in a high RME and low BMS. In contrast, only five projections are required for producing a reasonable CS reconstruction, as illustrated in \figref{phantom_5proj-b}. The phantom in the TVR-DART reconstruction, shown in \figref{phantom_5proj-c}, exhibits relatively consistent intensity but still contains several streaking artifacts and contrast variations. The CSHM reconstruction in \figref{phantom_5proj-d} achieves a high degree of uniform intensity while accurately reproducing the phantom's structure. This is indicated by a low RME of 0.042 and a high BMS of 0.974, although some contrast variations remain, particularly near the material's boundaries. As demonstrated in \figref{phantom_5proj-e}, the MIP RO method, using the selection of $\alpha = \beta$, yields the best reconstruction. There is almost no variation in the material's intensity, and the morphology and borders are reconstructed nearly perfectly when compared to the phantom's ground truth in \figref{dataset-a}. This is confirmed by an even lower RME of 0.017 and a higher BMS of 0.998 compared to the CSHM reconstruction. Only the runtime is the highest of all compared algorithms. However, as highlighted previously, parallelization may render this drawback neglectable. 

\begin{figure}[htbp]
        \scriptsize
        \begin{subfigure}[t]{0.19\textwidth}
            \centering
            \includegraphics[width=\textwidth]{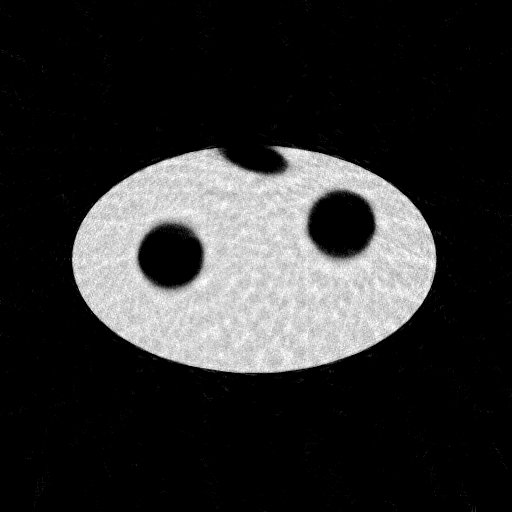}
            \caption{SIRT: 16 s, \\ RME: $0.190$, BMS: $0.773$}
            \label{phantom_20proj-a}
        \end{subfigure}
        \begin{subfigure}[t]{0.19\textwidth}
            \centering
            \includegraphics[width=\textwidth]{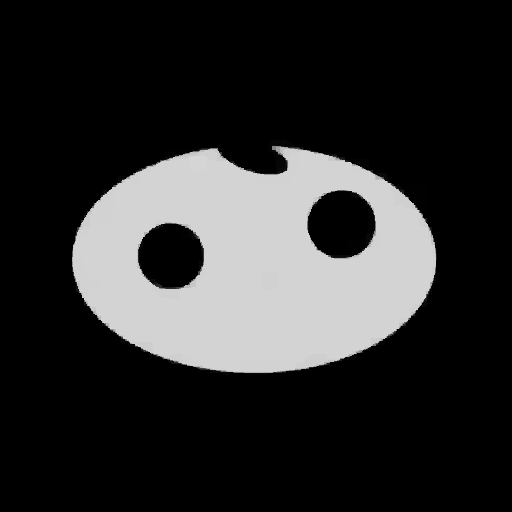}
            \caption{CS: 343 s,\\RME: $0.031$, BMS: $0.988$}
            \label{phantom_20proj-b}
        \end{subfigure}
        \begin{subfigure}[t]{0.19\textwidth}
            \centering
            \includegraphics[width=\textwidth]{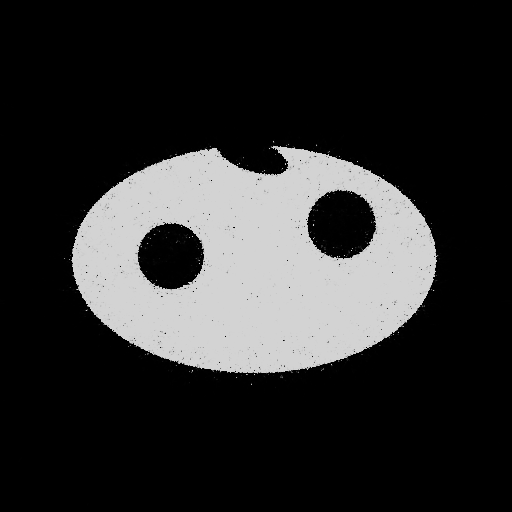}
            \caption{TVR-DART: 26 s,\\  RME: $0.043$, BMS: $0.990$}
            \label{phantom_20proj-c}
        \end{subfigure}
        \begin{subfigure}[t]{0.19\textwidth}
            \centering
            \includegraphics[width=\textwidth]{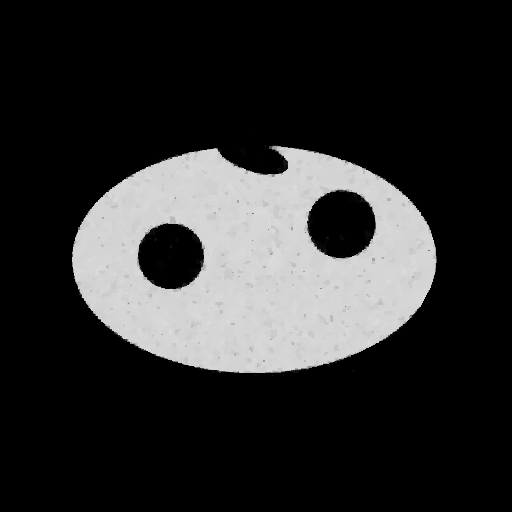}
            \caption{CSHM: 139 s,\\ RME: $ 0.060$, BMS: $0.795$}
            \label{phantom_20proj-d}
        \end{subfigure}
        \begin{subfigure}[t]{0.19\textwidth}
            \centering
            \includegraphics[width=\textwidth]{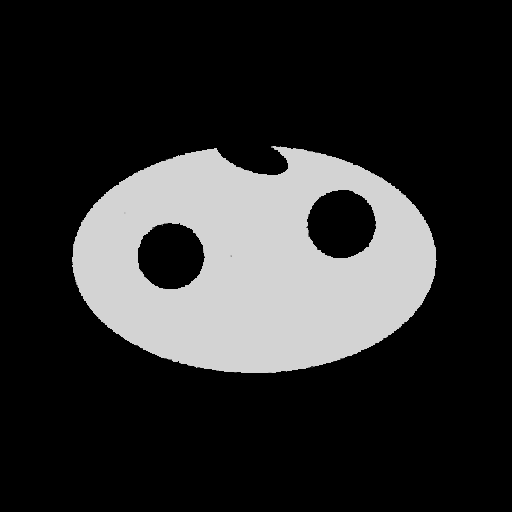}
            \caption{MIP RO: 687 s,\\ RME: 0.008, BMS: $0.999$}
            \label{phantom_20proj-e}
        \end{subfigure}
    \caption{\small Results for the simulated data with $512^2$ pixels and 20 projections -- $\lambda_{cs}=20000$, $\lambda_{TVR}=500$, $T=800$, $\alpha=\frac{1}{50}, \beta=\frac{1}{50}$.}
    \label{phantom_20proj}
\end{figure}
\FloatBarrier

The reconstructions generated using the same algorithms and parameters, but with an increased number of 20 projections, are presented in \figref{phantom_20proj}. With SIRT, the phantom's structure is now reconstructed quite faithfully; however, the interior still displays streaking artifacts and significant contrast variations, as shown in \figref{phantom_20proj-a}. The RME has decreased and the BMS has increased only slightly compared to the SIRT reconstruction with 5 projections in \figref{phantom_5proj-a}. The CS reconstruction in (\figref{phantom_20proj-b}) achieves a very high quality, exhibiting homogeneous intensity and accurately detailing structural features and sharp interfaces. A similar level of quality is observed for TVR-DART and CSHM reconstructions in \figref{phantom_20proj-c} and  \figref{phantom_20proj-d}. However, some speckling and contrast variations persist in the interior of the phantom caused by noise in the projection data, leading to a higher RME compared to the CS method. The TVR-DART approach produces a highly bimodal result with a high BMS, while the CSHM method shows greater contrast variations, resulting in a lower BMS. The MIP RO reconstruction displayed in \figref{phantom_20proj-e} achieves the highest fidelity overall. The interfaces are very sharp, and the material's intensity is nearly uniform with almost no speckles, leading to an even lower RME and a higher BMS compared to the MIP RO reconstruction with 5 projections in \figref{phantom_5proj-e}.

\subsection{Results for Zeolite Particle} 

\begin{figure}[htbp]      
        \centering
        \begin{subfigure}[t]{0.19\textwidth}
            \centering
            \includegraphics[width=\textwidth]{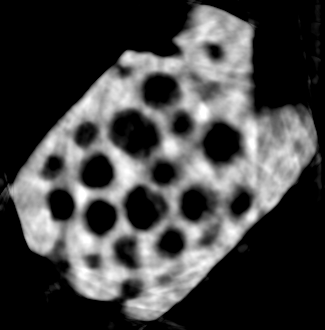}
            \caption{SIRT: 3 s, \\ RDC: $0.023$, BMS: $0.798$}
        \label{fig:zeolite_20proj-a}
        \end{subfigure}
        \begin{subfigure}[t]{0.19\textwidth}
            \centering
            \includegraphics[width=\textwidth]{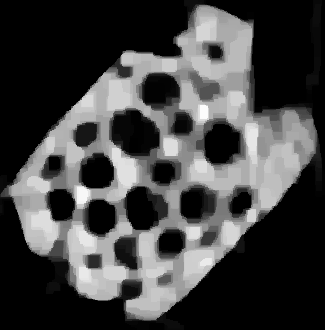}
            \caption{CS: 252 s,\\  RDC: $0.023$, BMS: $0.801$}
        \label{fig:zeolite_20proj-b}
        \end{subfigure}
        \begin{subfigure}[t]{0.19\textwidth}
            \centering
            \includegraphics[width=\textwidth]{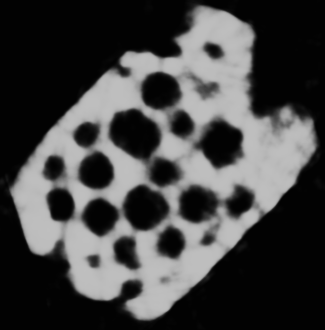}
            \caption{TVR-DART: 75 s,\\RDC: $0.070$, BMS: $0.876$}
        \label{fig:zeolite_20proj-c}
        \end{subfigure}
        \begin{subfigure}[t]{0.19\textwidth}
            \centering
            \includegraphics[width=\textwidth]{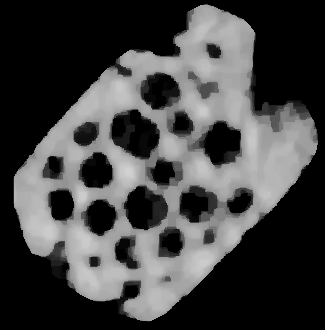}
            \caption{CSHM: 136 s,\\ RDC: $0.028$, BMS: $0.810$}
        \label{fig:zeolite_20proj-d}
        \end{subfigure}
        \begin{subfigure}[t]{0.19\textwidth}
            \centering
            \includegraphics[width=\textwidth]{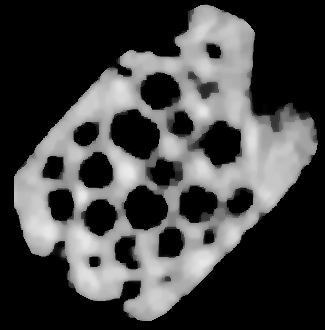}
            \caption{MIP: 6794 s,\\ RDC: $0.090$, BMS: $0.833$}
        \label{fig:zeolite_20proj-e}
        \end{subfigure}
    \caption{\small Results for the experimental electron tomography dataset of a zeolite particle with $512^2$ pixels and 20 projections -- $\lambda_{CS} =4000$, $\lambda_{TVR}=2000$ , $T=800$, $\alpha = \frac{1}{125}$, $\beta = \frac{1}{50}$.}
    \label{fig:zeolite_20proj}
\end{figure}

Next, we compare the reconstructions of an electron tomography dataset from a porous zeolite particle. 
This dataset consists of 20 and 30 projections taken from the tilt series shown in \figref{proj-showing-et1} with 180° tilt-angle range and equidistant tilt increments. One exemplary slice through the center of the zeolite particle is reconstructed with a pixel number of $512^2$, respectively. The results for $20$ projections are shown in \figref{fig:zeolite_20proj}. Below the reconstructions, we note the corresponding runtimes, RDCs and BMS values, along with the applied reconstruction parameters: $T$, $\alpha$, $\beta$ (MIP RO), $\lambda_{cs}$ and $\lambda_{TVR}$. \\
The SIRT reconstruction in \figref{fig:zeolite_20proj-a} suffers from significant streaking artifacts and contrast variations, which stem from the limited number of projections. Similarly, the CS reconstruction in \figref{fig:zeolite_20proj-b} displays intense contrast variations. However, these are confined to piecewise constant regions typical of CS, resulting in sharper edges than those seen in the SIRT reconstruction. The TVR-DART reconstruction shown in (\figref{fig:zeolite_20proj-c}) yields a much more homogeneous structure that seems to have high fidelity. However, it is important to note that the fine details reconstructed by TVR-DART may not accurately correspond to the ground truth, as evidenced by the absence of tiny channels between pores in the SIRT reconstruction with 180 projections (see \figref{dataset-b}). The CSHM reconstruction in \figref{fig:zeolite_20proj-d} demonstrates a much more uniform intensity compared to both SIRT and CS. This results in a clearer binary separation between zeolite material and pore space. Additionally, staircase artifacts characteristic of CS are reduced, and the interface is sharper than that in TVR-DART, with the lowest RDC value among all techniques used. \\


\begin{figure}[htbp]      
        \centering
        \begin{subfigure}[t]{0.24\textwidth}
            \centering
            \includegraphics[width=\textwidth]{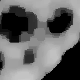}
            \caption{CSHM Region 1}
        \end{subfigure}
        \begin{subfigure}[t]{0.24\textwidth}
            \centering
            \includegraphics[width=\textwidth]{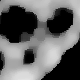}
            \caption{MIP RO Region 1}
        \end{subfigure}
        \begin{subfigure}[t]{0.24\textwidth}
            \centering
            \includegraphics[width=\textwidth]{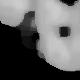}
            \caption{CSHM Region 2}
        \end{subfigure}
        \begin{subfigure}[t]{0.24\textwidth}
            \centering
            \includegraphics[width=\textwidth]{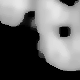}
            \caption{MIP RO Region 2 }
        \end{subfigure}
    \caption{\small CSHM vs. MIP RO: local comparison of two regions of interest for the zeolite particle with $512^2$ pixels and 20 projections -- $\lambda_{CS} =4000$, $\lambda_{TVR}=2000$, $T=800$, $\alpha = \frac{1}{125}$, $\beta = \frac{1}{50}$.}
    \label{fig:CSHMvsMIP RO-20proj}
\end{figure}

In comparison to the other reconstructions, the MIP RO method in \figref{fig:zeolite_20proj-e} strikes an excellent balance between minimizing contrast variations and maintaining detail fidelity (see Supplementary Video 3 for an animated video of slices through the full reconstruction of the zeolite particle at a pixel number of 256²). It produces the sharpest interfaces overall, and staircase artifacts within the zeolite are completely absent, unlike in the CS and CSHM reconstructions. MIP RO results in a slightly higher overall intensity for the zeolite compared to CSHM. This is because washed-out edge intensities and less steep gradient interfaces are confined to steeper boundaries. Consequently, these removed intensities are redistributed into a smaller number of pixels assigned to the zeolite material, resulting in a higher overall intensity. This can be clearly seen in the zoomed-in ROIs displayed in \figref{fig:CSHMvsMIP RO-20proj}. MIP RO effectively reconstructs sharp interfaces and reduces the presence of washed-out regions, which are somewhat visible in other reconstruction methods, including the SIRT reconstruction with 180 projections in \figref{dataset-b}. Remaining contrast variations in materials with intrinsically constant chemical composition and density, along with washed-out regions in experimental datasets, can often be attributed to imperfect experimental conditions, such as Bragg contrast, beam broadening, and slight misalignment in the tilt series. These inconsistencies also account for the artifacts and flaws observed in all reconstructions and contribute to the non-zero RDC values. MIP RO addresses these inconsistencies by allowing for solutions that may not result in the best RDC but prioritize solution that consider prior knowledge, in particular the materials exhibiting constant density and sharp interfaces. For the experimental zeolite dataset, selecting $\alpha < \beta$ yielded the best reconstructions for MIP RO. Instead of producing a strictly binary image, as seen with $\alpha = \beta$, the reconstructed intensities are gently pushed toward binary values, enabling greater grayscale variation in the reconstruction. For further details, we refer to Section \ref{subsubsec:MIP RO-weightings}. \\

\FloatBarrier

\subsubsection{Threshold for Edge Decision}

\begin{figure}[htbp]      
    \centering
     \begin{subfigure}[b]{0.32\textwidth}
            \centering
            \includegraphics[width=\textwidth]{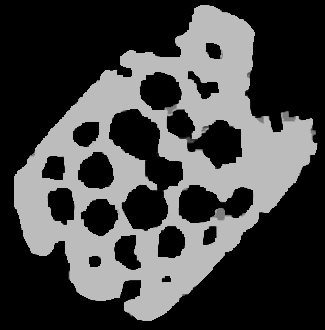}
            \caption{$T$ = 400, MC: 39135, BMS: 0.993}
            \label{TinfluenceArea-a}
        \end{subfigure}
        \begin{subfigure}[b]{0.32\textwidth}
            \centering
            \includegraphics[width=\textwidth]{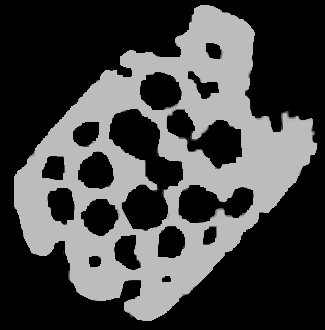}
            \caption{$T$ = 600, MC: 39236, BMS: 0.993}
            \label{TinfluenceArea-b}
        \end{subfigure}
        \begin{subfigure}[b]{0.32\textwidth}
            \centering
            \includegraphics[width=\textwidth]{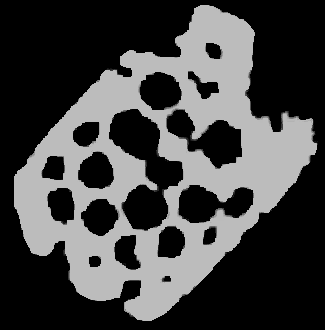}
            \caption{$T$ = 800, MC: 39508, BMS: 0.991}
            \label{TinfluenceArea-c}
        \end{subfigure}
    \caption{\small Results for the experimental electron tomography dataset of a zeolite particle of varying $T$ with $\alpha=\frac{1}{50}, \beta=\frac{1}{50}$ with $512^2$ pixels and 20 projections.}
    \label{TinfluenceArea}
\end{figure}

\figref{TinfluenceArea} illustrates how the parameter $T$ from equation \eqref{MIP_obj_a} (see also Section \ref{subsubsec:threshold}) 
affects the MIP RO reconstruction results by determining whether an edge is sharp enough to be detected. When $T$ is low, sub-regions are more likely to be classified as edges. Conversely, a higher $T$ tends to categorize a sub-region as a region of uniform density. Moreover, a low $T$ can suppress weaker edges, typically resulting in lower material coverage (MC) and larger pore spaces. This effect is exemplified in \figref{TinfluenceArea-a}, where $T$ = 400. The increased sensitivity to edge detection at low $T$ may lead to  pore interface regions, where multiple edge lines are detected. This can result in rougher pore surfaces and washed-out intensities, as seen in the regions of intermediate grey intensity in \figref{TinfluenceArea-a}. In contrast, when $T$ is set to higher values, as shown in \figref{TinfluenceArea-b} ($T$ = 600) and \figref{TinfluenceArea-c} ($T$ = 800), edge detection occurs even for weaker edges (i.e., lower Sobel values). This results in a greater number of pixels being assigned the material's intensity, leading to a higher MC value. Specifically for the zeolite structure, a higher $T$ leads to more uniform edge detection at the boundaries where multiple edge lines (i.e. local Sobel maxima) are closely clustered together. \\

\subsubsection{Weightings in Objective of MIP RO}
\label{subsubsec:MIP RO-weightings}

\begin{figure}[htbp]      
    \centering
     \begin{subfigure}[b]{0.32\textwidth}
            \centering
            \includegraphics[width=\textwidth]{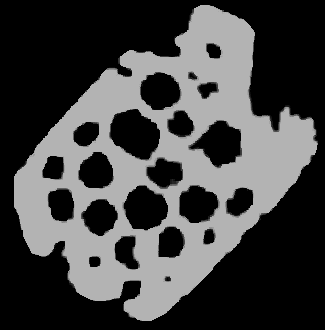}
            \caption{$\beta=\frac{1}{125}$, BMS: $0.985$,  runtime: 1205 s}
            \label{fig:beta-BMS-1-a}
        \end{subfigure}
        \begin{subfigure}[b]{0.32\textwidth}
        \label{fig: b variation 1 by 50}
            \centering
            \includegraphics[width=\textwidth]{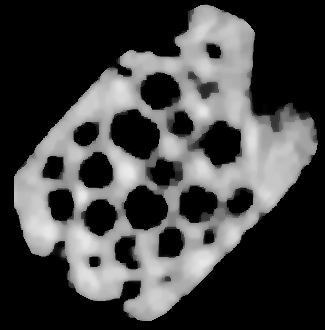}
            \caption{$\beta=\frac{1}{50}$, BMS: $0.833$, runtime: 8553 s}
            \label{fig:beta-BMS-1-b}
        \end{subfigure}
        \begin{subfigure}[b]{0.32\textwidth}
        \label{fig: b variation 1 by 25}
            \centering
            \includegraphics[width=\textwidth]{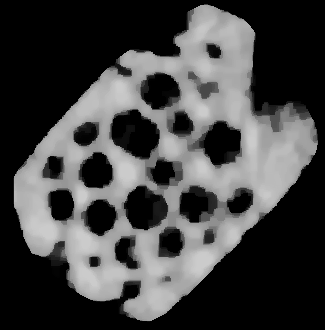}
            \caption{$\beta=\frac{1}{15}$, BMS: $0.824$, runtime: 3359 s}
            \label{fig:beta-BMS-1-c}
        \end{subfigure}
    \caption{\small MIP RO impact of varying $\beta$ with T = 800, $\alpha=\frac{1}{125}$, $512^2$ pixels with 20 projections.}
    \label{fig:beta-BMS-1}
\end{figure}

As explained in Section \ref{subsubsec:alpha,beta}, the parameter $\beta$ can be adjusted to control the degree of bimodality in the reconstructed intensities.\\
When $\alpha = \beta$, the reconstructed image is compelled to display a high degree of bimodality, as illustrated in \figref{fig:beta-BMS-1-a} for the zeolite structure, which shows a BMS close to 1. This setting also results in the fastest runtime due to the linear nature of the objective function. This approach is effective for idealized datasets, as seen in \figref{phantom_5proj} and \figref{phantom_20proj}, which feature a phantom with added noise.\\
In contrast, when $\alpha < \beta$, the reconstructed intensities are only gently nudged toward binary values (i.e., representing either empty space or material), allowing for more grayscale variation in the reconstruction. This is demonstrated in \figref{fig:beta-BMS-1-b} and \figref{fig:beta-BMS-1-c}, which result in lower BMS values. However, due to the non-convex nature of the objective function, this scenario leads to slower running times compared to the linear case.\\
For objects made of a single material phase, a binary reconstruction is typically preferred. However, in experimental datasets, permitting grayscale variations may be advantageous and may prevent forced binarization. While \figref{fig:beta-BMS-1-a} shows clearer edges and a more uniform zeolite structure compared to the SIRT reconstruction using 180 projections in \figref{dataset-b}, some pore walls appear to be misidentified, and the overall pore space seems over-segmented (i.e., too large). Inconsistencies in the acquired raw data -- such as nonlinear contrast contributions from Bragg scattering, beam broadening, or misalignment in the tilt series -- often necessitate allowing for some variability in grayscale values.\\
\figref{fig:beta-BMS-1-b} and \figref{fig:beta-BMS-1-c} exemplify this, as both allow for slight grayscale variations (indicated by a lower BMS) that enable the reconstructions to detect thinner pore walls that were missed in \figref{fig:beta-BMS-1-a}.  Nonetheless, the pore-zeolite interface remains very distinct across all reconstructions, successfully preventing washed-out regions.\\


\begin{figure}[htbp] 
        \centering
        \begin{subfigure}[t]{0.24\textwidth}
            \centering
            \includegraphics[width=\textwidth]{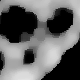}
            \caption{$\beta=\frac{1}{50}$ Region 1}
        \end{subfigure}
        \begin{subfigure}[t]{0.24\textwidth}
            \centering
            \includegraphics[width=\textwidth]{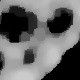}
            \caption{$\beta=\frac{1}{15}$ Region 1}
        \end{subfigure}
        \begin{subfigure}[t]{0.24\textwidth}
            \centering
            \includegraphics[width=\textwidth]{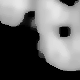}
            \caption{$\beta=\frac{1}{50}$  Region 2}
        \end{subfigure}
        \begin{subfigure}[t]{0.24\textwidth}
            \centering
            \includegraphics[width=\textwidth]{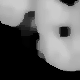}
            \caption{$\beta=\frac{1}{15}$  Region 2 }
        \end{subfigure}
    \caption{\small Displaying impact of $\beta=\frac{1}{50}$ vs. $\beta=\frac{1}{15}$ on two regions within the zeolite structure, reconstructed with $512^2$ pixels, $T=800$, $\alpha = \frac{1}{125}$ and 20 projections.}
    \label{fig:beta-BMS-2}
\end{figure}

\figref{fig:beta-BMS-2} compares selected regions of interest from \figref{fig:beta-BMS-1} at higher magnification, focusing on the sharpness of their interfaces. The reconstruction with a smaller value of $\beta=\frac{1}{50}$ is closer to $\alpha$ and displays a sharper interface with fewer grayscale variations (i.e. higher BMS) compared to the result with $\beta=\frac{1}{15}$.\\

\FloatBarrier

\FloatBarrier
\subsection{Results for Copper  Microlattice}

\begin{figure}[htbp]      
        \centering
        \begin{subfigure}[t]{0.19\textwidth}
            \centering
            \includegraphics[width=\textwidth]{images_v1/3_Cu_microlattice/sirt_slice702_512_40_angles_lambda3.png}
            \caption{SIRT: 6 s, \\ RDC: $0.072$, BMS: $0.640$}
            \label{fig:Cu-Lattice_1-a}
        \end{subfigure}
        \begin{subfigure}[t]{0.19\textwidth}
            \centering
            \includegraphics[width=\textwidth]{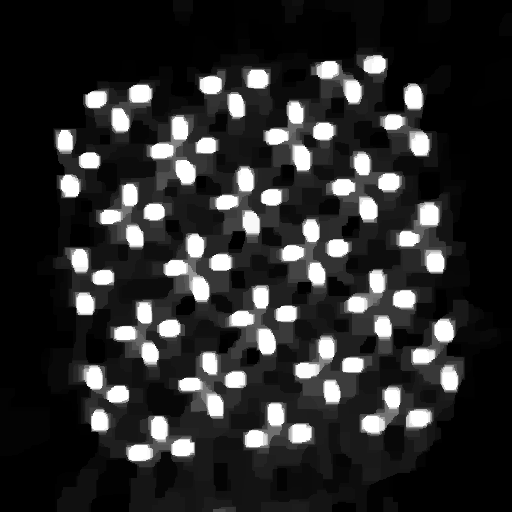}
            \caption{CS: 984 s,\\RDC: $0.069$, BMS: $0.539$}
            \label{fig:Cu-Lattice_1-b}
        \end{subfigure}
        \begin{subfigure}[t]{0.19\textwidth}
            \centering
            \includegraphics[width=\textwidth]{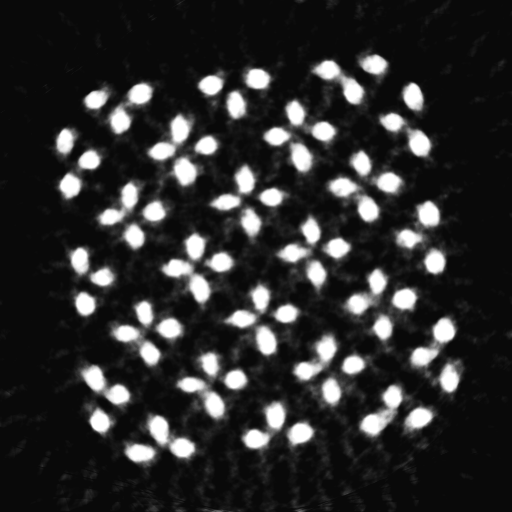}
            \caption{TVR-DART: 12 s,\\  RDC: $0.088$, BMS: $0.681$}
            \label{fig:Cu-Lattice_1-c}
        \end{subfigure}
        \begin{subfigure}[t]{0.19\textwidth}
            \centering
            \includegraphics[width=\textwidth]{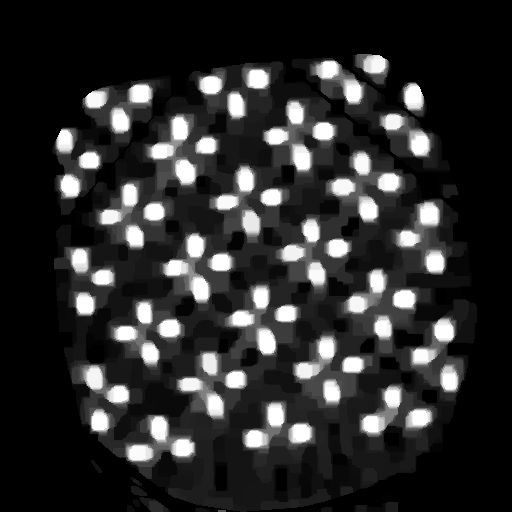}
            \caption{CSHM: 622 s,\\ RDC: $0.058$, BMS: $0.521$ }
            \label{fig:Cu-Lattice_1-d}
        \end{subfigure}
        \begin{subfigure}[t]{0.19\textwidth}
            \centering
            \includegraphics[width=\textwidth]{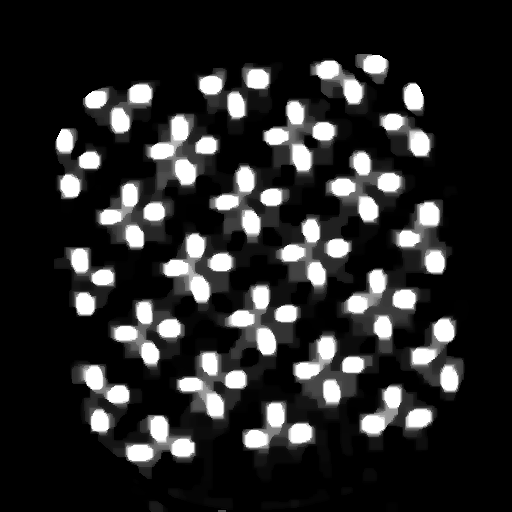}
            \caption{MIP RO: 12838 s,\\ RDC: $0.126$, BMS: $0.763$}
            \label{fig:Cu-Lattice_1-e}
        \end{subfigure}
    \caption{\small Results for the experimental nano-CT dataset of a Cu microlattice with $512^2$ pixels, 40 projections, $\lambda_{CS}=3, \lambda_{TVR}=7500, \alpha=\frac{1}{125}, \beta=\frac{1}{25}, T=800$.}
    \label{fig:Cu-Lattice_1}
\end{figure}

We now compare the reconstructions of an absorption contrast nano-CT dataset obtained from a copper microlattice structure. This dataset comprises 40 projections taken from the tilt series shown in \figref{proj-showing-et2}, covering a 180° tilt-angle range with equidistant tilt increments. One exemplary slice through the center of the copper microlattice is reconstructed with $512^2$ pixels. The results are presented in \figref{fig:Cu-Lattice_1}. Below the reconstructions, we provide the corresponding runtimes, RDC values, and BMS values, along with the applied reconstruction parameters: $\lambda_{cs}$, $\lambda_{TVR}$,  $\alpha$, $\beta$ and $T$.

\FloatBarrier

Due to the limited number of projections, the SIRT reconstruction shown in \figref{fig:Cu-Lattice_1-a} exhibits the most prominent artifacts, both inside and outside the structure. In contrast, these artifacts are significantly reduced in the CS, TVR-DART, CSHM, and MIP RO reconstructions presented in \figref{fig:Cu-Lattice_1-b}-\figref{fig:Cu-Lattice_1-e}. The CS reconstruction results in piecewise constant regions that represent the sharply separated copper struts of consistent intensity. However, some incorrectly allocated intensities still surround the struts. In \figref{fig:Cu-Lattice_1-c}, the TVR-DART reconstruction appears to offer high fidelity, but it still contains a noticeable amount of noise and finer artifacts. The CSHM reconstruction is quite similar to CS but achieves the lowest RDC value among the methods presented.

MIP RO provides the cleanest reconstruction, featuring the most distinct copper struts and sharpest edges. It minimizes artifacts outside the struts compared to the other methods, while also showing the fewest artifacts present between the copper struts. This is reflected in the highest overall BMS, although it comes with an increased RDC and the longest runtime if parallelization is not utilized.

Similar to the electron tomography dataset of the zeolite particle, MIP RO prioritizes achieving the sharpest and cleanest reconstruction of the copper microlattice rather than the lowest RDC. This deviation from the lowest RDC can be justified by considering that the sample consists of mostly homogeneous copper struts with sharp boundaries and that the nano-CT tilt series may contain inconsistencies such as noise and nonlinear contrast contributions due to the relatively high X-ray attenuation of copper for the employed X-ray energy.

\begin{figure}[htbp]      
        \centering
        \begin{subfigure}[t]{0.28\textwidth}
            \includegraphics[width=\textwidth]{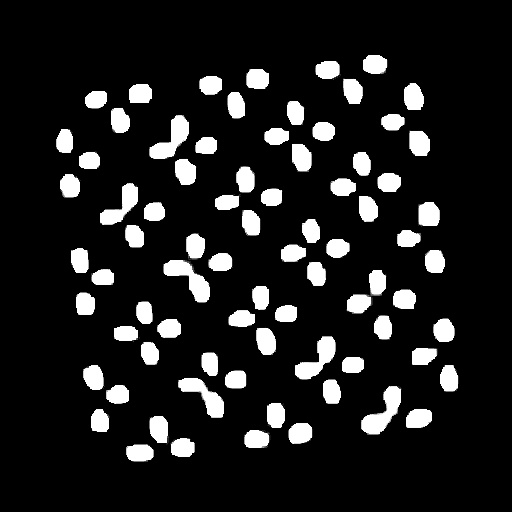}
            \caption{MIP RO: 1049 s,\\ $\alpha=\frac{1}{50},\beta=\frac{1}{50}$  \\RDC: $0.318$, BMS: $0.990$ }
            \label{fig:Cu-Lattice_2-a}
        \end{subfigure}
        \begin{subfigure}[t]{0.28\textwidth}
            \includegraphics[width=\textwidth]{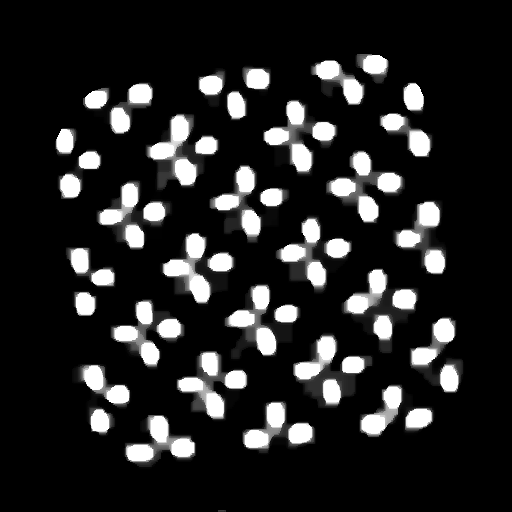}
            \caption{MIP RO: 7778 s,\\ $\alpha=\frac{1}{125},\beta=\frac{1}{50}$  \\RDC: $0.204$, BMS: $0.883$}
            \label{fig:Cu-Lattice_2-b}
        \end{subfigure}
        \begin{subfigure}[t]{0.28\textwidth}
            \includegraphics[width=\textwidth]{images_v1/3_Cu_microlattice/MIPRO_512_702_40__T800_a0.008_b0.04_R12838_num_sr260100.png}
            \caption{MIP RO: 12838 s,\\ $\alpha=\frac{1}{125},\beta=\frac{1}{25}$  \\RDC: $0.126$, BMS: $0.763$ }
            \label{fig:Cu-Lattice_2-c}
        \end{subfigure}
    \caption{\small Influence of different values for $\alpha$ and $\beta$ on MIP RO results for the experimental nano-CT dataset of a Cu microlattice with $512^2$ pixels, 40 projections, and T = 800.}
    \label{fig:Cu-Lattice_2}
\end{figure}
\FloatBarrier

\figref{fig:Cu-Lattice_2} illustrates the effect of different values for $\alpha$ and $\beta$ on the MIP RO results. When $\alpha = \beta$ (\figref{fig:Cu-Lattice_2-a}), the resulting image is the most binary, also being reflected by the highest BMS. Similar to the findings for the zeolite particle in \figref{fig:beta-BMS-1}, allowing for grayscale variations in experimental datasets can be beneficial and may prevent forced binarization. Consequently, selecting $\alpha < \beta$ results in qualitatively improved outcomes, as depicted in  \figref{fig:Cu-Lattice_2-b} and \figref{fig:Cu-Lattice_2-c}, which is also reflected in a lower RDC.

\subsection{Missing Wedge}

\begin{figure}[htbp]      
        \centering
        \begin{subfigure}[t]{0.19\textwidth}
            \centering
            \includegraphics[width=\textwidth]{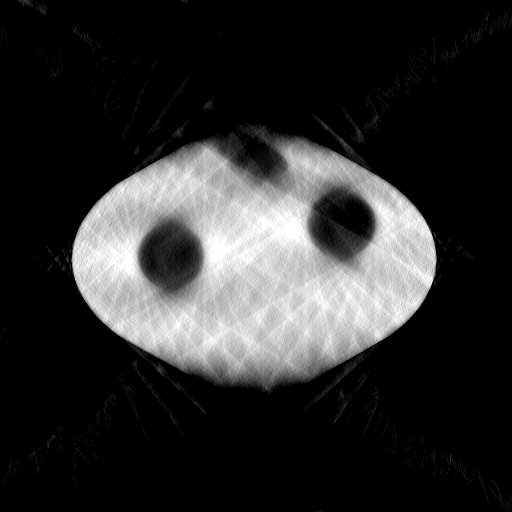}
            \caption{SIRT: 3 s, \\ RME: $0.506$, BMS: $0.731$}
        \label{fig:missingwedge-a}
        \end{subfigure}
        \begin{subfigure}[t]{0.19\textwidth}
            \centering
            \includegraphics[width=\textwidth]{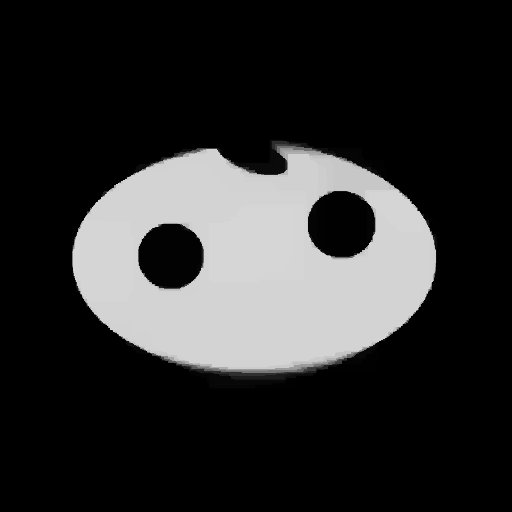}
            \caption{CS: 256 s,\\RME: $0.038$, BMS: $0.970$}
        \label{fig:missingwedge-b}
        \end{subfigure}
        \begin{subfigure}[t]{0.19\textwidth}
            \centering
            \includegraphics[width=\textwidth]{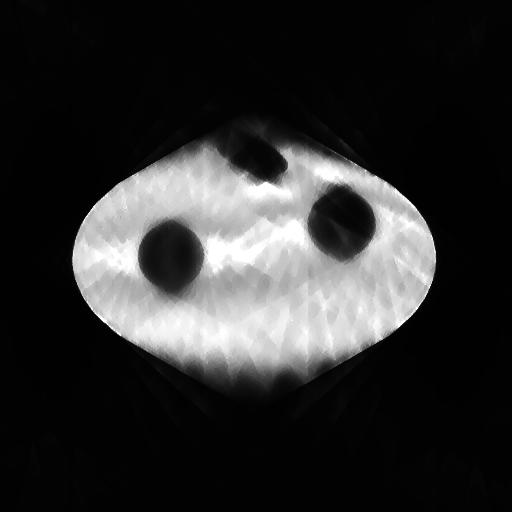}
            \caption{TVR-DART: 12 s,\\  RME: $0.236$, BMS: $0.746$}
        \label{fig:missingwedge-c}
        \end{subfigure}
        \begin{subfigure}[t]{0.19\textwidth}
            \centering
            \includegraphics[width=\textwidth]{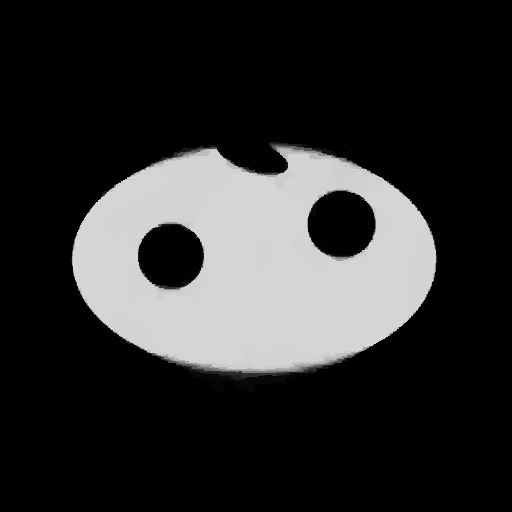}
            \caption{CSHM: 66 s,\\ RME: $0.032$, BMS: $ 0.978$ }
        \label{fig:missingwedge-d}
        \end{subfigure}
        \begin{subfigure}[t]{0.19\textwidth}
            \centering
            \includegraphics[width=\textwidth]{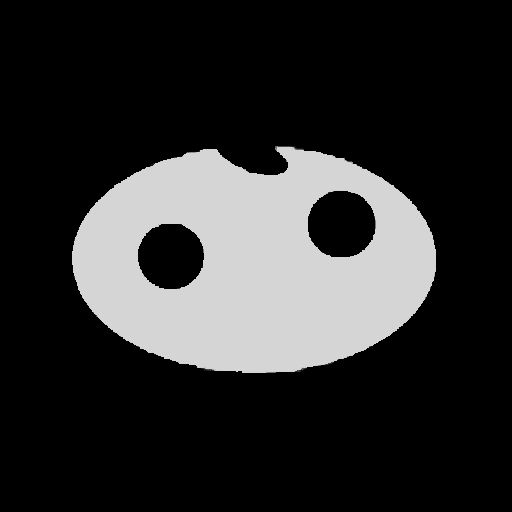}
            \caption{MIP RO: 656 s,\\ RME: $0.010$, BMS: $0.999$}
        \label{fig:missingwedge-e}
        \end{subfigure}

        \begin{subfigure}[t]{0.19\textwidth}
            \centering
            \includegraphics[width=\textwidth]{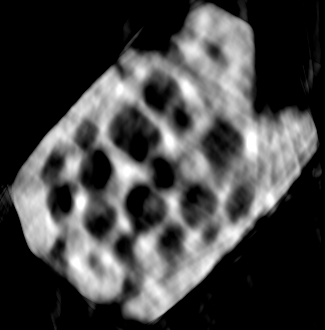}
            \caption{SIRT: 4 s, \\ RDC: $0.146$, BMS: $0.780$}
        \label{fig:missingwedge-f}
        \end{subfigure}
        \begin{subfigure}[t]{0.19\textwidth}
            \centering
            \includegraphics[width=\textwidth]{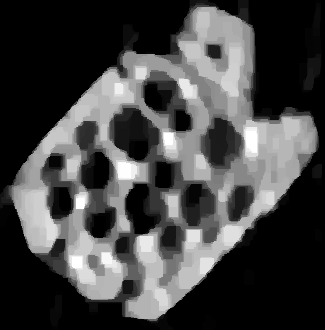}
            \caption{CS: 360s,\\RDC: $0.020$, BMS: $0.788$}
        \label{fig:missingwedge-g}
        \end{subfigure}
        \begin{subfigure}[t]{0.19\textwidth}
            \centering
            \includegraphics[width=\textwidth]{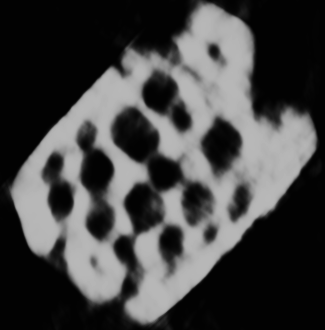}
            \caption{TVR-DART: 12 s,\\  RDC: $0.080$, BMS: $0.856$}
        \label{fig:missingwedge-h}
        \end{subfigure}
        \begin{subfigure}[t]{0.19\textwidth}
            \centering
            \includegraphics[width=\textwidth]{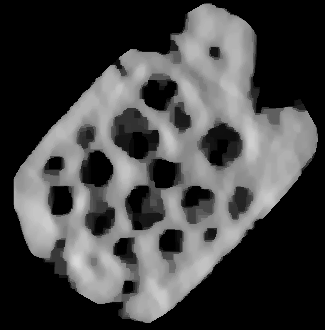}
            \caption{CSHM: 109 s,\\ RDC: $0.030$, BMS: $0.797$ }
        \label{fig:missingwedge-i}
        \end{subfigure}
        \begin{subfigure}[t]{0.19\textwidth}
            \centering
            \includegraphics[width=\textwidth]{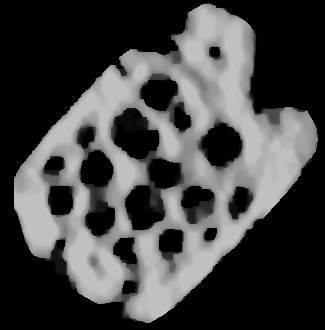}
            \caption{MIP RO: 5499 s,\\ RDC: $0.324$, BMS: $0.872$}
        \label{fig:missingwedge-j}
        \end{subfigure}
    \caption{\small Results for missing wedge of 60° with $512^2$ pixels and 11 projections for the simulated dataset (a)–(d) and 21 projections for the zeolite particle dataset (e)–(h). $\lambda_{TVR} = 100, T=800, \alpha=\frac{1}{50}, \beta=\frac{1}{50}$ for the simulated, $\lambda_{TVR} = 5000, T=800, \alpha=\frac{1}{125}, \beta=\frac{1}{50}$ for the zeolite data.}
    \label{fig:missingwedge}
\end{figure}

\figref{fig:missingwedge} presents reconstructions of the phantom structure (upper row) and the zeolite particle (lower row) with a missing wedge of 60°. The phantom, characterized by its relatively simple geometric structure and constant intensity, is well reconstructed by CS, CSHM, and MIP RO, even with just 11 projections. Only the TVR-DART method struggles to find an optimal solution. Among these methods, MIP RO in \figref{fig:missingwedge-e} delivers the best reconstruction, exhibiting minimal contrast variations and very sharp edges—performing even better than both CS and CSHM. This performance is reflected in its lowest RME and highest BMS. In the missing-wedge scenario for the zeolite particle, the SIRT reconstruction displays typical strong elongation and streaking artifacts, as shown in \figref{fig:missingwedge-f}. Although the other reconstruction methods reduce these missing-wedge artifacts, CS still suffers from significant contrast variations and staircase artifacts (\figref{fig:missingwedge-g}). Even though the artificial vertical elongation of the particle's outline is mostly mitigated, several pores are incorrectly merged that would remain distinct in a complete reconstruction. TVR-DART achieves a solution with very uniform zeolite intensities and fine details. However, the missing wedge causes a misleading merging of pores along the vertical direction, as observed in \figref{fig:missingwedge-h}. The CSHM reconstruction results in a zeolite structure with homogeneous intensity (\figref{fig:missingwedge-i}).  There is little apparent elongation of the pores, and most pore borders are reconstructed correctly when compared to the reconstructions without the missing wedge (\figref{fig:zeolite_20proj}). The MIP RO reconstruction in \figref{fig:missingwedge-j} achieves even sharper edges and fewer incorrectly assigned intensities within the pores compared to CSHM, resulting in overall better contrast between the pores and the zeolite structure. While some pore walls are reconstructed with lower intensity compared to CSHM, they are still maintained.

\section{Conclusion}\label{sec:conclusion}

In the previous section, we demonstrated the performance of our MIP RO approach applied to one simulated dataset and two experimental datasets obtained from two different nanotomography techniques, namely 360°-ET and nano-CT. The results indicate that this method effectively meets the objectives of improving the CSHM approach, particularly regarding the enhancement of sharp interfaces. MIP RO was compared to CSHM and three other benchmark algorithms SIRT, CS, and TVR-DART. We explored various problem settings, including scenarios with a missing wedge.

For the simulated data set, MIP RO achieved superior reconstruction quality and the lowest relative mean error compared to all other techniques. Specifically, both MIP RO as well as the integrated approach from Section \ref{sec:framework_integrated} successfully incorporated DNN constraints into an underlying MIP model to reconstruct images, which yield sharper edges and more homogeneous reconstruction of the material. Consequently, an overall improved fidelity albeit increased runtimes was achieved. However, it is important to note, that in contrast to CSHM and other compressed sensing techniques including the integrated approach, MIP RO allows for significant parallelization, which could be exploited in future research. 

For the experimental datasets, MIP RO effectively reduced artifacts both outside and within the reconstructed samples, while also increasing edge sharpness. In terms of parameter selection, setting $\alpha = \beta$ in \eqref{eq:MIP-RO} results in a solution with a high degree of binarization. In contrast, choosing $\alpha < \beta$ allows for deviations from the intended homogeneous intensity value, which is particularly advantageous for experimental datasets prone to acquisition inconsistencies. Fine-tuning these parameters enables either the generation of sharp, binary images or, in cases of more inconsistent raw data, a balanced image that retains intensity variations while still featuring sharp edges. However, it is worth noting that the non-convex nature of the objective function may lead to increased runtimes.

Similar to all the compared algorithms, the reconstructions obtained from MIP RO in the context of the missing-wedge problem still exhibit typical artifacts. In particular, artifacts such as elongation, as well as the merging of features, which would appear separate in reconstructions without a missing wedge, are still present. However, MIP RO demonstrates the least merging of pores and elongation while achieving the best overall contrast between the pores and the material, which remains very homogeneous.

In summary, MIP RO is a powerful enhancement to the already competitive CSHM algorithm for addressing small- to medium-sized image reconstruction problems involving homogeneous samples exhibiting features confined by sharp edges. Rather than primarily focusing on achieving the lowest raw data coverage (RDC), MIP RO prioritizes the enhancement of sharp edges and uniform material contrast, which is particularly advantageous for inconsistent and undersampled raw data, including scenarios involving missing wedges.

Looking ahead, an extension of this method to handle samples composed of multiple density values is a promising area for future research and would call for a more sophisticated training procedure for the neural network. In addition, approximating further sample properties and imaging physics via DNN have the potential to improve MIP RO even further.

Finally, we point out that our approach is not limited to imaging applications; the concept can also be applied to other problems where knowledge extracted from data is incorporated into an optimization model. 

\section*{Data availability statement}
The Python source code of the implemented algorithms and the experimenntal tilt series raw data is publicly available on Zenodo under \\
\href{https://doi.org/10.5281/zenodo.16587222}{https://doi.org/10.5281/zenodo.16587222} and \\ 
\href{https://doi.org/10.5281/zenodo.10283560}{https://doi.org/10.5281/zenodo.10283560}.

\section*{Acknowledgement}
The paper is funded by the Deutsche Forschungsgemeinschaft (DFG, German Research Foundation) -
Project-ID 416229255 - SFB 1411.
We thank Sung-Gyu Kang (Gyeongsang
National University, Republic of Korea) and Rajaprakash Ramachandramoorthy (MPIE Düsseldorf, Germany) for the provision of the copper microlattice sample, which had been developed within the framework of the 2023 ERC Starting Grant AMMicro (grant number 101078619), and Wilhelm Schwieger (CRT, FAU Erlangen-Nürnberg) for providing the zeolite sample. The authors are grateful to Sebastian Kreuz for support with using the CSHM algorithm and to Erdmann Spiecker (IMN $\&$ CENEM, FAU Erlangen-Nürnberg) for fruitful discussions on image reconstruction.
The authors gratefully acknowledge the scientific support and HPC resources provided by the Erlangen National High Performance Computing Center (NHR@FAU) of the FAU. The hardware is funded by the DFG.




\bibliographystyle{elsarticle-harv} 
\bibliography{literature}

\end{document}